\documentclass{article}

\usepackage[nonatbib, preprint]{neurips_2026}

\usepackage[utf8]{inputenc}
\usepackage[T1]{fontenc}
\usepackage{hyperref}
\usepackage{url}
\usepackage{booktabs}
\usepackage{amsmath,amssymb,amsfonts}
\usepackage{graphicx}
\usepackage{subcaption}
\usepackage[export]{adjustbox}
\usepackage{array}
\usepackage{tabularx}
\usepackage{microtype}
\usepackage{xcolor}
\usepackage{placeins}

\usepackage[
    backend=biber,
    style=numeric-comp,
    sorting=none,
    giveninits=true,
    maxbibnames=6,
    minbibnames=1,
    doi=false,
    url=false,
    isbn=false,
    eprint=true,
]{biblatex}
\AtEveryBibitem{%
  \clearfield{abstract}%
  \clearfield{file}%
  \clearfield{issn}%
  \clearfield{keywords}%
  \clearfield{langid}%
  \clearfield{month}%
  \clearfield{note}%
  \clearfield{pubstate}%
  \clearfield{shorttitle}%
  \clearfield{urldate}%
  \clearlist{language}%
}

\title{How Far Do Simple Transformations Translate Across Text Embedding Models?}

\author{%
  Sid Ali Hamideche \\
  Orange \\
  \texttt{sidali.hamideche@orange.com}
  \And
  Louis-Adrien Dufrène\\
  Orange \\
  \texttt{louisadrien.dufrene@orange.com}
  \And
  Quentin Lampin \\
  Orange \\
  \texttt{quentin.lampin@orange.com}
  \And
  Guillaume Larue \\
  Orange \\
  \texttt{guillaume.larue@orange.com}
}

\begin{document}

\maketitle
\begin{abstract}
	We investigate whether simple transformations can translate representations across heterogeneous text embedding models. Understanding how independently trained models organize semantic information is an enabler for AI-to-AI latent communication without decoding into human-readable text. Focusing on lightweight translators such as linear mappings, we test the literature hypothesis of latent universality in a realistic text setting beyond simplified benchmarks. Across nine embedding models differing in architecture, pooling strategy, and training objective, we evaluate compatibility using CKA, downstream transfer, fidelity, and retrieval. Simple translators recover meaningful shared structure and support transfer for some compatible pairs, but fail sharply for others. Compatibility depends jointly on architecture, training objective, pooling, and data distribution. Overall, the results show that heterogeneous embedding spaces are not universally related by simple mappings as often suggested in some literature.
\end{abstract}

\thanks{This research was funded in part by the SNS JU 6GARROW project under the EU’s Horizon program Grant Agreement No 101192194, in collaboration with ROK.}

\section{Introduction}

Text embedding models are increasingly used as interchangeable infrastructure for retrieval, classification, and multimodal systems. This trend raises a fundamental question: when two models process the same text, how similar are their representations, and can a simple transformation (e.g., a linear map) relate them?

This question matters for several reasons. Simple translation between the representations of two models could make pretrained components easier to reuse with other models, enable a more modular ecosystem of embedding-based tools and pipelines, and enable stitching scenarios where a head trained on one model reuses translated features from another \cite{bansalRevisitingModelStitching2021b, chenTransferringLinearFeatures2025b}. It also probes whether independently trained systems share geometric organization \cite{huhPlatonicRepresentationHypothesis2024a, kornblithSimilarityNeuralNetwork2019a}, and allows easier interpretation of how models represent concepts and relations. Additionally, one of our target applications is agent-to-agent communication beyond natural language, so understanding translation limits is foundational for a longer-term direction: AI agents communicating via latent spaces rather than decoded text, avoiding information loss from projecting continuous states onto discrete tokens \cite{zouLatentCollaborationMultiAgent2025a, duEnablingAgentsCommunicate2026b, zhuSurveyLatentReasoning2025}. The present study characterizes where simple transformations work and where they fail, challenging the notion of universal latent compatibility often stated in the literature \cite{huhWhatMakesImageNet2016, lencUnderstandingImageRepresentations2015a, csiszarikSimilarityMatchingNeural2021a, smithOfflineBilingualWord2017, moschellaRelativeRepresentationsEnable2023b, lahnerDirectAlignmentLatent2024a, maiorcaLatentSpaceTranslation2023a, gorbettCharacterizingLinearAlignment2026b}.

In this work, we first evaluate a constrained regime that admits closed-form solutions and strong interpretability: direct linear maps from small anchor sets \cite{maiorcaLatentSpaceTranslation2023a}, relative representations \cite{moschellaRelativeRepresentationsEnable2023b}, and inverse relative projection \cite{maiorcaLatentSpaceTranslation2024b}. These methods not only state that some shared structure exists but also state that a simple, anchor-based alignment is enough to recover it. We then include a more flexible linear map trained on a larger paired corpus, as an empirical approximate upper bound on what linear transformations can achieve with more supervision \cite{lahnerDirectAlignmentLatent2024a}.

Our study extends prior work by evaluating a heterogeneous collection of models varying in dimension, tokenizer, pooling, normalization, architecture, and training objective, and by combining complementary diagnostics, CKA, downstream transfer, fidelity, and retrieval, rather than relying on a single metric.


\paragraph{Contributions.}
\begin{itemize}
	\item A systematic empirical study of simple translation approaches across heterogeneous text embedding models.
	\item Multi-diagnostic evaluation covering geometric similarity, retrieval, and downstream transfer, going beyond simple benchmarks such as classification models trained on MNIST or CIFAR.
	\item Evidence that simple transformations are useful but strongly pair-dependent, with architectural and training explanations for compatibility patterns.
\end{itemize}

\section{Related Work and Translation Methods}
\label{sec:related}

We organize the related literature around the two families of simple translators we evaluate, defining each method in the context of the work that proposed it. Throughout, $X \in \mathbb{R}^{d_A \times m}$ and $Y \in \mathbb{R}^{d_B \times m}$ denote matrices whose columns are paired embeddings of $m$ inputs in models $A$ and $B$.

\paragraph{Direct linear maps.}
A long line of work has shown that representations of independently trained models can be related, to a non-trivial extent, by an unconstrained linear map. Lenc and Vedaldi \cite{lencUnderstandingImageRepresentations2015a} first demonstrated this for image models; Bansal et al.\ \cite{bansalRevisitingModelStitching2021b} revisited the idea as model stitching, using the linear map itself as a similarity probe. More recently, L\"ahner and Moeller \cite{lahnerDirectAlignmentLatent2024a} reported that unconstrained linear transforms outperform isometric ones, and Gorbett and Jana \cite{gorbettCharacterizingLinearAlignment2026b}, along with Chen et al.\ \cite{chenTransferringLinearFeatures2025b}, showed that affine maps largely preserve downstream performance across LLM representations. Concretely, given paired embeddings we seek $T \in \mathbb{R}^{d_B \times d_A}$ minimizing $\| Y - T X \|_F^2$. The closed-form least-squares solution
\[
	T^* \;=\; Y X^+,
\]
which we refer to as \emph{Linear (pinv)}, requires only a small set of paired anchors and underlies the formalization of Maiorca et al.\ \cite{maiorcaLatentSpaceTranslation2023a}. When more paired data is available, the same objective can be optimized by stochastic gradient descent on mini-batches of pairs, which we call \emph{Linear (SGD)} \cite{lahnerDirectAlignmentLatent2024a} and use as an empirical upper bound on what an unconstrained linear translator can achieve.

\paragraph{Anchor-based and relative representations.}
A complementary line of work translates by re-expressing each embedding in terms of its relation to a shared anchor set. Moschella et al.\ \cite{moschellaRelativeRepresentationsEnable2023b} introduced \emph{relative representations} (RR), which replace an embedding $x$ with the vector of cosine similarities to $m$ anchors $\{a_1, \dots, a_m\}$ in the same model,
\[
	r(x) \;=\; \big[\, \cos(x, a_1),\; \cos(x, a_2),\; \dots,\; \cos(x, a_m) \,\big]^\top \;\in\; \mathbb{R}^m,
\]
producing a coordinate system whose dimensions carry direct semantic readings (each entry measures similarity to a named anchor concept) and that is approximately invariant to latent isometries. This invariance enables zero-shot stitching, but it constrains the downstream head to be trained on the relative representation rather than on a native embedding space, narrowing the applicability of the method. Inverse Relative Projection (IRP) \cite{maiorcaLatentSpaceTranslation2024b} lifts this constraint by learning a map from the relative space back into a target model's usable embedding space; Cannistraci et al.\ \cite{cannistraciBricksBridgesProduct2024a} further incorporate product invariances to strengthen latent communication. Section~\ref{sec:local-structure} makes precise the sense in which RR, IRP, and \emph{Linear (pinv)} are variations on the same anchor-based template.

\paragraph{Representational similarity and latent communication.}
Beyond the translators themselves, a parallel literature quantifies how similar two representation spaces are without explicitly aligning them. CKA \cite{kornblithSimilarityNeuralNetwork2019a} and related measures \cite{csiszarikSimilarityMatchingNeural2021a} score geometric alignment across models, and the Platonic Representation Hypothesis \cite{huhPlatonicRepresentationHypothesis2024a} conjectures that such similarity grows as models scale. Recent work on latent agent communication \cite{zouLatentCollaborationMultiAgent2025a, duEnablingAgentsCommunicate2026b, zhuSurveyLatentReasoning2025} motivates cross-model translation as the substrate for efficient AI-to-AI interfaces, the longer-term application that frames the present study.

\section{Experimental Setup}

We evaluate nine text embedding models spanning encoder-only, T5-style encoders, and a causal-backbone model (Table~\ref{tab:models-list}), intentionally diverse in dimension, tokenizer, pooling, normalization, and training objective.

\begin{table}[ht]
	\caption{Embedding models included in the study.}
	\label{tab:models-list}
	\centering
\resizebox{\textwidth}{!}{
	\begin{tabular}{|l|c|c|c|>{\centering\arraybackslash}p{2.5cm}|>{\centering\arraybackslash}p{2.5cm}|>{\centering\arraybackslash}p{1.2cm}|}
		\hline
		Model Name            & Dimension & Tokenizer               & Architecture       & Pooling     & Normalization by default & Max Length \\
		\hline
		Qwen3-Embedding-4B    & 2560      & Qwen2TokenizerFast      & Qwen3ForCausalLM   & last token  & Yes                      & 40960      \\
		\hline
		e5-large-v2           & 1024      & BertTokenizerFast       & BertModel          & mean tokens & Yes                      & 512        \\
		\hline
		mxbai-embed-large-v1  & 1024      & BertTokenizerFast       & BertModel          & cls token   & No                       & 512        \\
		\hline
		all-roberta-large-v1  & 1024      & RobertaTokenizerFast    & RobertaForMaskedLM & mean tokens & Yes                      & 256        \\
		\hline
		bge-m3                & 1024      & XLMRobertaTokenizerFast & XLMRobertaModel    & cls token   & Yes                      & 8192       \\
		\hline
		gtr-t5-large          & 768       & T5TokenizerFast         & T5EncoderModel     & mean tokens & Yes                      & 512        \\
		\hline
		instructor-xl         & 768       & T5TokenizerFast         & T5EncoderModel     & mean tokens & Yes                      & 512        \\
		\hline
		all-mpnet-base-v2     & 768       & MPNetTokenizerFast      & MPNetForMaskedLM   & mean tokens & Yes                      & 384        \\
		\hline
		nomic-embed-text-v1.5 & 768       & BertTokenizerFast       & NomicBertModel     & mean tokens & No                       & 8192       \\
		\hline
	\end{tabular}
}

\end{table}

As for the data, for anchor-based methods we use the Basic English 850 word list, compact, human-inspectable, and semantically broad which can be motivated by using human expertise to select anchors. We also use 850 sentences from Wikitext-103, selected by computing the most distant sentences from the others (average across models), in terms of cosine similarity of their embeddings. We found no consistent difference in performance between them, nor significant gains from more elaborate anchor selection, so we keep the pipeline simple. However, we admit that we might not have pushed this step as far as the authors \cite{moschellaRelativeRepresentationsEnable2023b, maiorcaLatentSpaceTranslation2023a, maiorcaLatentSpaceTranslation2024b} of the original papers did. However, the focus of this study is not anchor selection but rather to show whether there exists a universal transformation between the embeddings of any pair of models.
As for \emph{Linear (SGD)} we use Wikitext-103 sentence pairs to train a linear map for each model pair. This can be seen as an approximate upper bound on what a linear map can achieve and compensates for not having further refined the anchor selection step for the anchor-based methods.

Evaluation uses four diagnostics: (1) CKA for geometric similarity, (2) downstream classification transfer, (3) representation fidelity (cosine similarity and normalized error norm on held-out words), and (4) $k$-NN retrieval. The downstream benchmark includes SST-2, AG News, emotion, CoLA, DBpedia, IMDB, MRPC, and QQP \cite{muennighoffMTEBMassiveText2023, enevoldsenMMTEBMassiveMultilingual2025}, evaluated with a fixed RBF-kernel SVM. Hyperparameter tuning and model selection to maximize per-task performance were not the focus of this work; thus, they were kept minimal. We compare translators under a controlled interface rather than maximize per-model scores.

\section{Local Semantic Linear Structure}
\label{sec:local-structure}

Before evaluating cross-model transfer, it is worth recalling why a linear translator is a reasonable starting point at all. Word embeddings have long been observed to encode certain semantic relations as approximately linear directions: the canonical \textit{king}--\textit{queen}--\textit{man}--\textit{woman} parallelogram of Mikolov et al.\ \cite{mikolovEfficientEstimationWord2013, mikolovDistributedRepresentationsWords2013}, the bilingual lexicon induction of Lample et al.\ \cite{lampleWordTranslationParallel2018}, and the formal account of Allen and Hospedales \cite{allenAnalogiesExplainedUnderstanding2019} all rest on the same premise: that meaning composes, locally, as linear combinations of a small set of reference points. Replicating their analogy-style reconstruction across our nine heterogeneous text encoders, we observe that this property does carry over: sign patterns and coarse weightings of the decomposition remain stable across architectures, even though precise coefficients depend on each model's normalization and scale. We defer the empirical analysis to Appendix~\ref{app:analogy} and use the rest of this section to make explicit why this property is exactly the geometric prerequisite that makes anchor-based linear translation well defined.

Let $X \in \mathbb{R}^{d_A \times m}$ and $Y \in \mathbb{R}^{d_B \times m}$ collect the embeddings of $m$ shared anchors in models $A$ and $B$. Reconstructing a source vector $x \in \mathbb{R}^{d_A}$ as a least-squares combination of the source-side anchors yields coefficients
\[
	c^*(x) \;=\; \arg\min_{c \in \mathbb{R}^m}\; \| x - X c \|_2^2 \;=\; X^+ x,
\]
which are nothing more than the multi-anchor generalization of the analogy coefficients above: \textit{king}~$\approx \alpha\,$\textit{queen}~$+\,\beta\,$\textit{man}~$+\,\gamma\,$\textit{woman} is the same equation written with three anchors instead of $m$. Transporting the coefficients to the target side, on the assumption that the local linear geometry is preserved, gives
\[
	\hat{y} \;=\; Y\,c^*(x) \;=\; Y X^+ x \;=\; T_{A \rightarrow B}\, x,
	\qquad T_{A \rightarrow B} \;=\; Y X^+,
\]
which is exactly the closed-form \emph{Linear (pinv)} estimator of Maiorca et al.\ \cite{maiorcaLatentSpaceTranslation2023a}. Coefficient-based translation through shared anchors and direct pseudo-inverse alignment are therefore not two different algorithms but two readings of the same operation: one expresses the source point in an anchor basis, the other applies that same expression on the target side.

Two consequences follow. First, \emph{Linear (pinv)} can only succeed to the extent that the anchor-basis decomposition $c^*(x)$ is approximately preserved when crossing from $A$ to $B$, exactly the property the analogy experiment probes, and exactly the property that becomes increasingly fragile as the basis grows or as the two models differ in normalization (Appendix~\ref{app:analogy}). Second, the other anchor-based methods we evaluate are variations on the same theme. Relative representations \cite{moschellaRelativeRepresentationsEnable2023b} replace the affine coefficients $X^+ x$ with the cosine-similarity vector to the anchors, trading reconstruction fidelity for invariance to latent isometries; inverse relative projection \cite{maiorcaLatentSpaceTranslation2024b} then learns the analogue of $Y$ from the relative space back to a usable target embedding. All three methods inherit the same implicit assumption, that the local linear geometry around a small set of shared anchors is shared across models, which the rest of the paper puts to the test.

\section{Global Alignment and Transfer}

\subsection{Global Similarity Through CKA}

We compute linear CKA \cite{kornblithSimilarityNeuralNetwork2019a} over anchor embeddings: $\mathrm{CKA}(X,Y) = \|Y^\top X\|_F^2 / (\|X^\top X\|_F \cdot \|Y^\top Y\|_F)$. This metric is often used in literature to compare between the embeddings and is invariant to orthogonal transformations and isotropic scaling.

\begin{figure}[ht]
	\centering
	\includegraphics[width=0.6\textwidth]{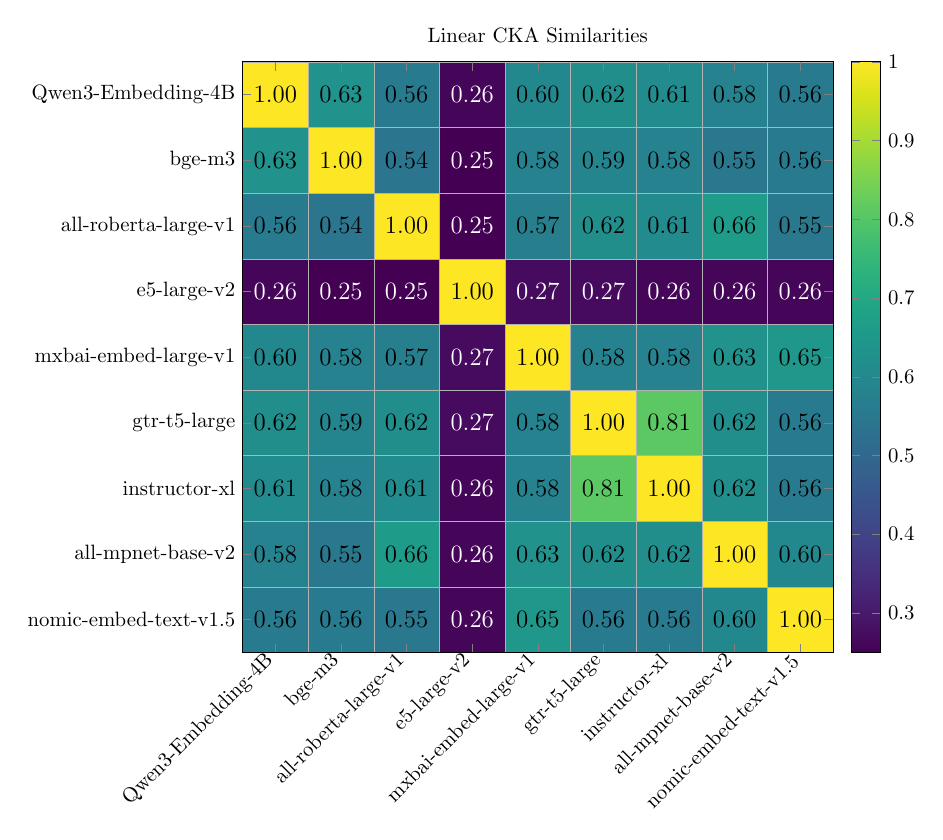}
	\caption{CKA similarity over Basic English anchors.}
	\label{fig:paper-cka-similarity-matrix}
\end{figure}

Figure~\ref{fig:paper-cka-similarity-matrix} reveals a structured, non-uniform compatibility pattern. \texttt{gtr-t5-large} and \texttt{instructor-xl} form the most similar pair: both use T5 encoder backbones with mean pooling, and contrastive fine-tuning on similar datasets, their near-identical training pipeline yields similar geometry. \texttt{all-mpnet-base-v2} and \texttt{all-roberta-large-v1} show moderate CKA with each other despite different base dimension. \texttt{e5-large-v2} is a pronounced outlier, despite using BERT with mean pooling, properties shared by several high-CKA models. Its pretraining, as stated by its authors \cite{wangTextEmbeddingsWeaklySupervised2024}, produces a geometry that diverges from the rest (cosine similarity in the range [0.7, 1] instead of [-1, 1], affecting a lot of angle-sensitive metrics like CKA). This observation already suggests that the universality up to a simple transformation might be too strong. It only needs one model to diverge in one aspect to question the universality. Note, however, that CKA is not invariant to arbitrary linear transformations.

\subsection{Downstream Task Transfer}

\begin{figure}[ht]
	\centering
	\begin{subfigure}{0.98\textwidth}
		\caption{AG News}
		\begin{subfigure}[t]{0.304\textwidth}
			\adjincludegraphics[width=\textwidth, trim={0 0 {0.1\width} 0}, clip]{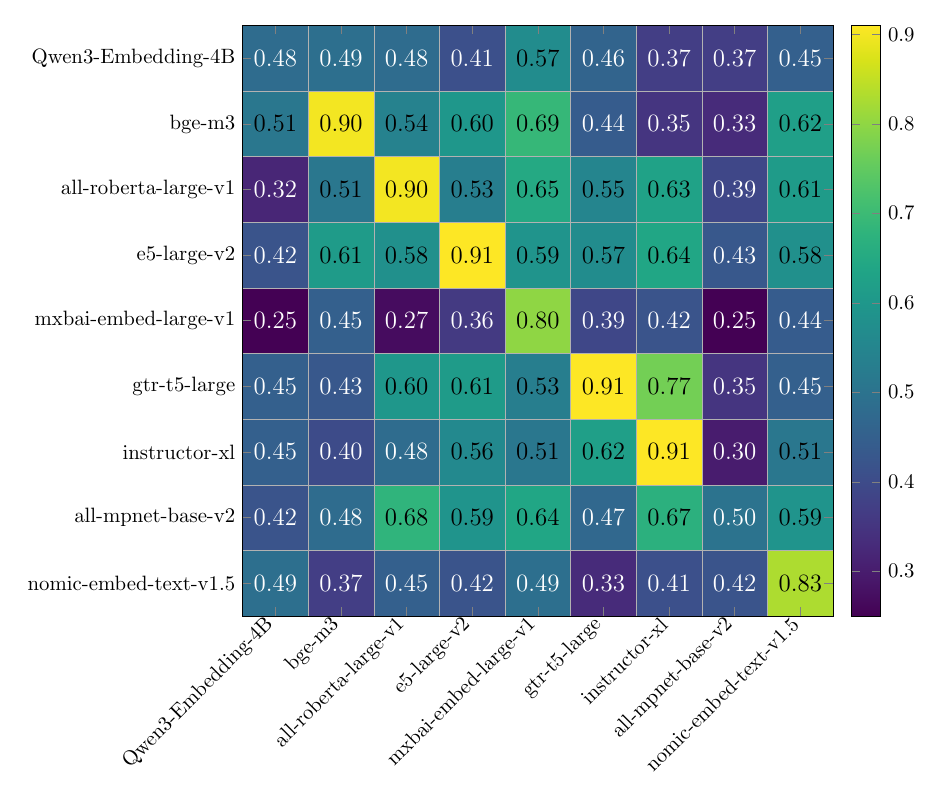}
			\caption*{Linear (pinv)}
		\end{subfigure}%
		\begin{subfigure}[t]{0.22\textwidth}
			\adjincludegraphics[width=\textwidth, trim={{0.25\width} 0 {0.1\width} 0}, clip]{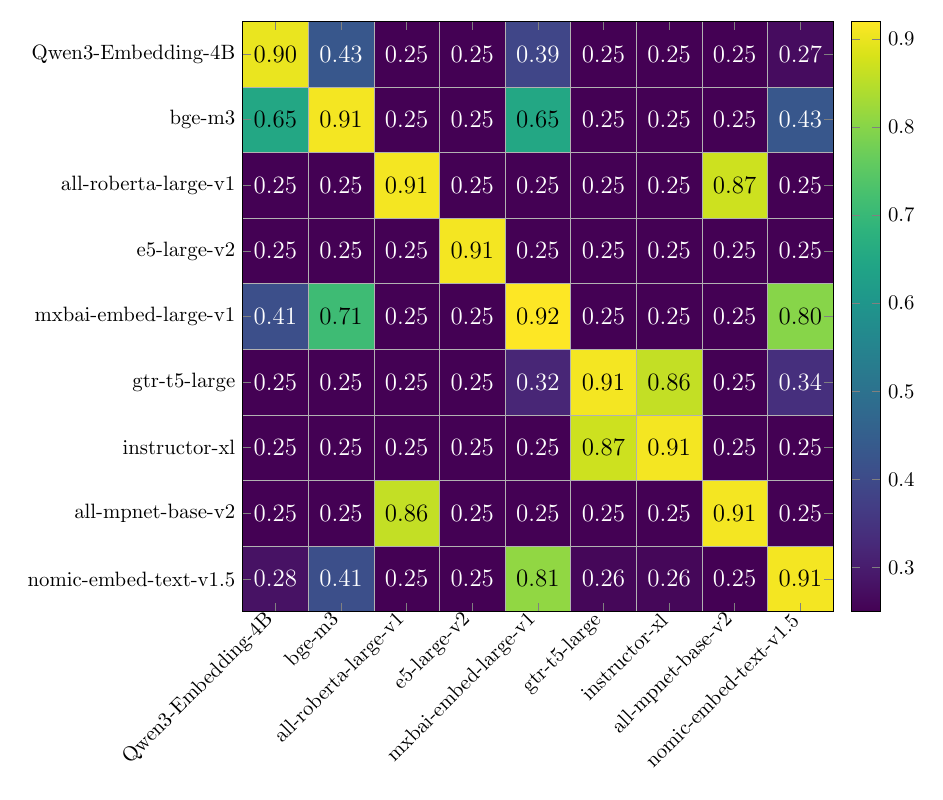}
			\caption*{RR}
		\end{subfigure}%
		\begin{subfigure}[t]{0.22\textwidth}
			\adjincludegraphics[width=\textwidth, trim={{0.25\width} 0 {0.1\width} 0}, clip]{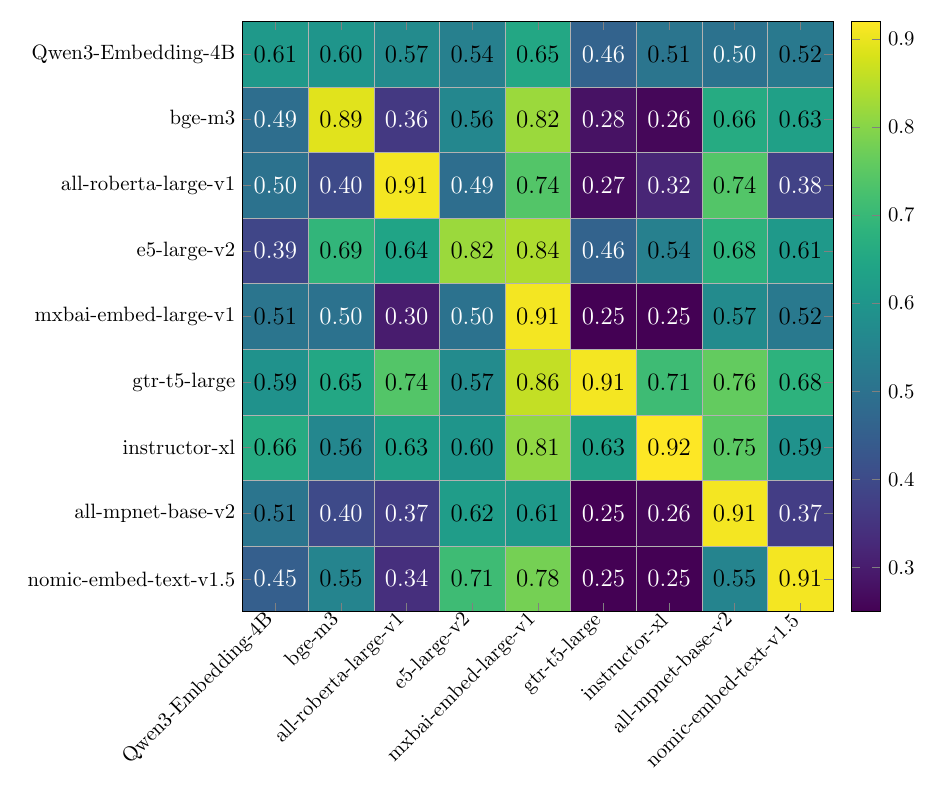}
			\caption*{IRP}
		\end{subfigure}%
		\begin{subfigure}[t]{0.256\textwidth}
			\adjincludegraphics[width=\textwidth, trim={{0.25\width} 0 0 0}, clip]{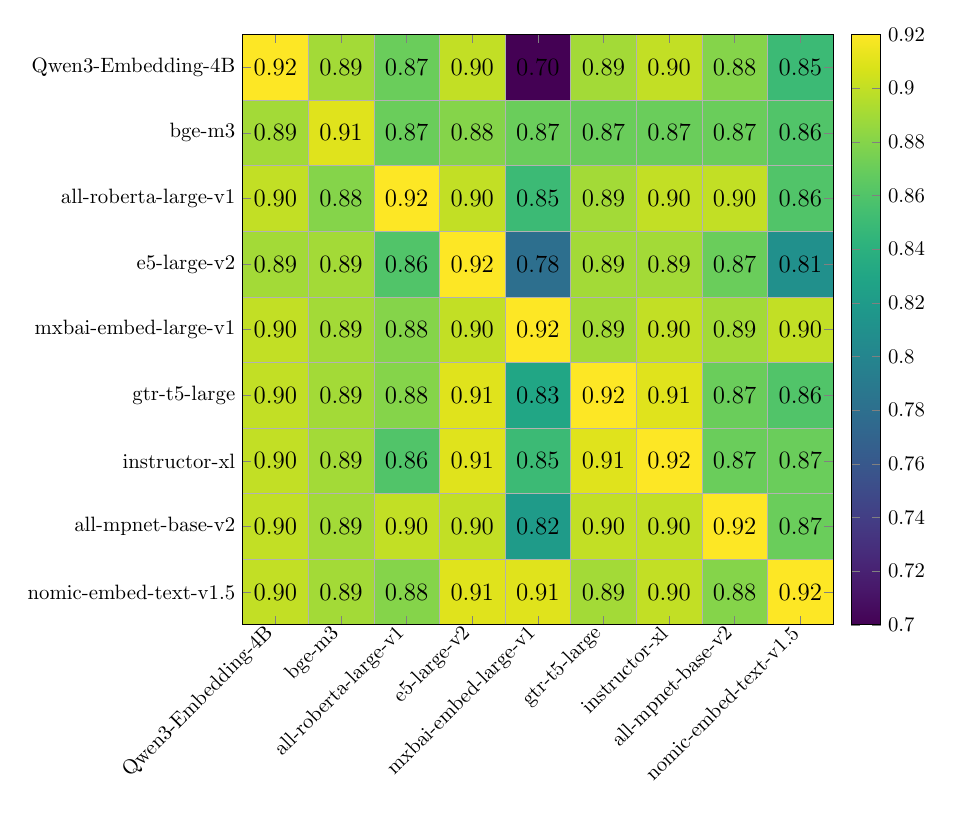}
			\caption*{Linear (SGD)}
		\end{subfigure}%
	\end{subfigure}
	\hfill
	\begin{subfigure}{0.98\textwidth}
		\caption{DBpedia}
		\begin{subfigure}[t]{0.304\textwidth}
			\adjincludegraphics[width=\textwidth, trim={0 0 {0.1\width} 0}, clip]{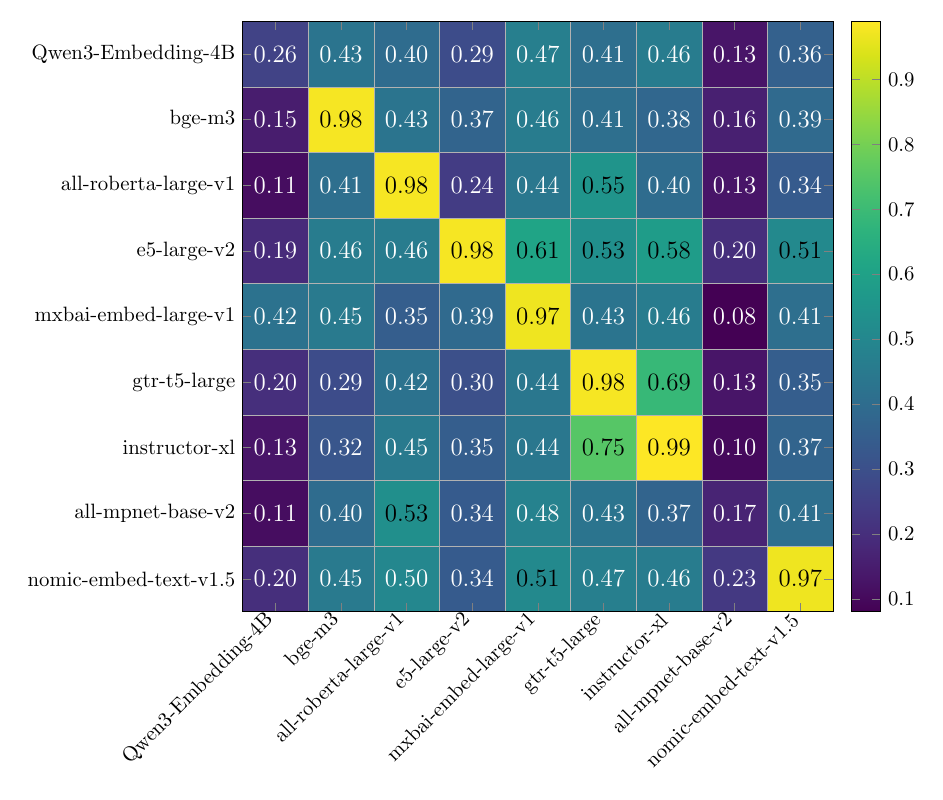}
			\caption*{Linear (pinv)}
		\end{subfigure}%
		\begin{subfigure}[t]{0.22\textwidth}
			\adjincludegraphics[width=\textwidth, trim={{0.25\width} 0 {0.1\width} 0}, clip]{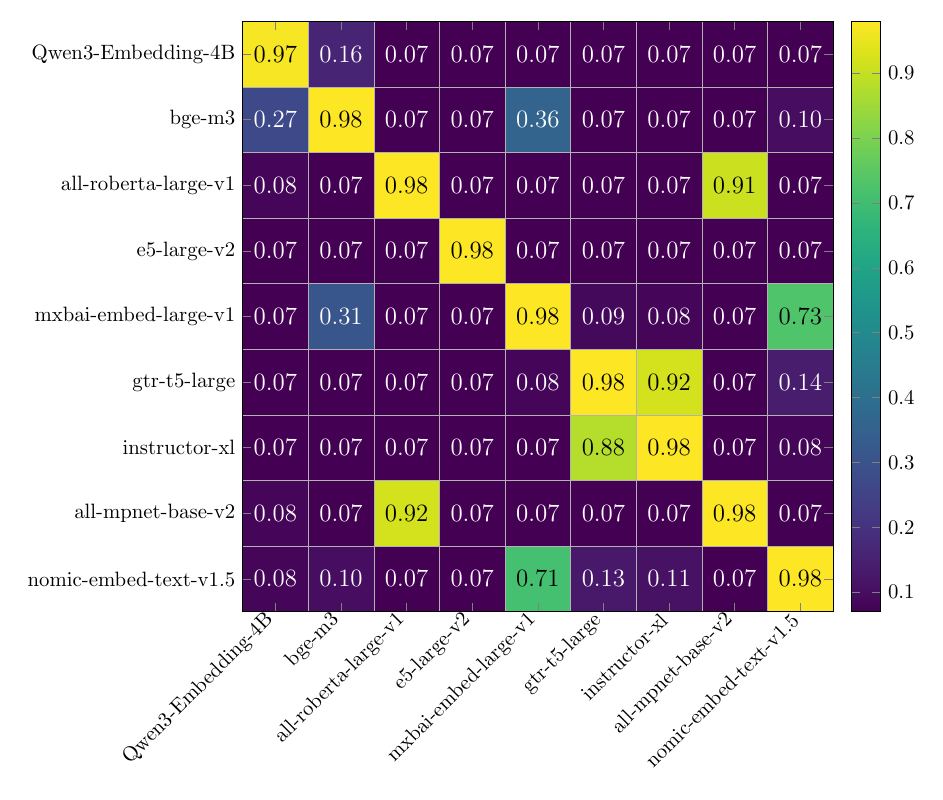}
			\caption*{RR}
		\end{subfigure}%
		\begin{subfigure}[t]{0.22\textwidth}
			\adjincludegraphics[width=\textwidth, trim={{0.25\width} 0 {0.1\width} 0}, clip]{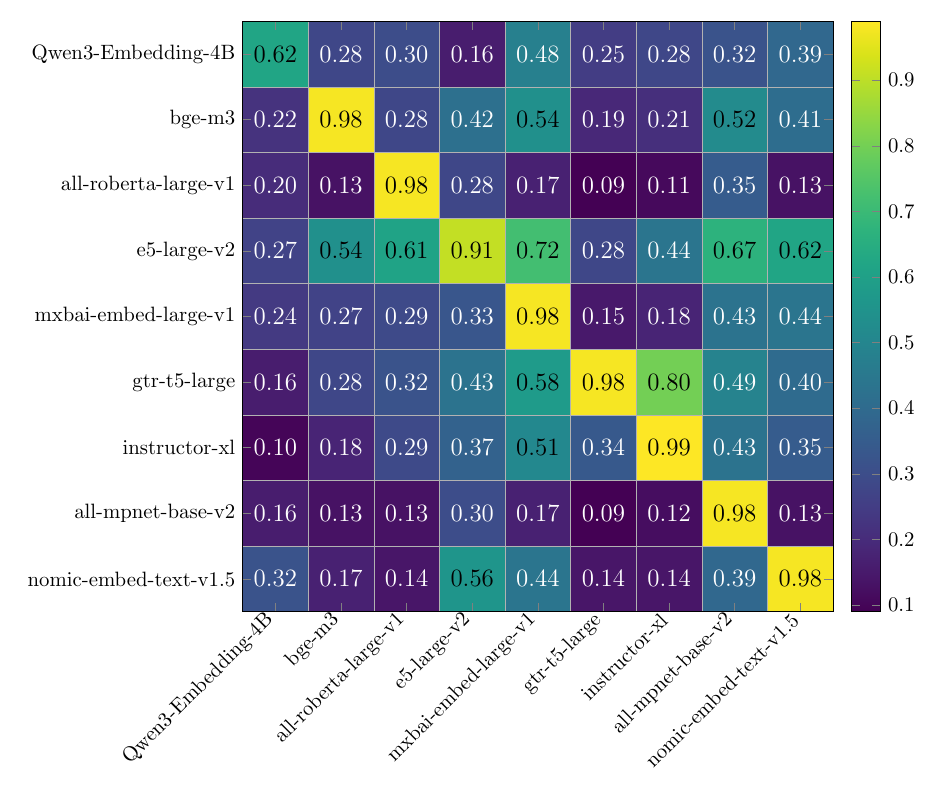}
			\caption*{IRP}
		\end{subfigure}%
		\begin{subfigure}[t]{0.256\textwidth}
			\adjincludegraphics[width=\textwidth, trim={{0.25\width} 0 0 0}, clip]{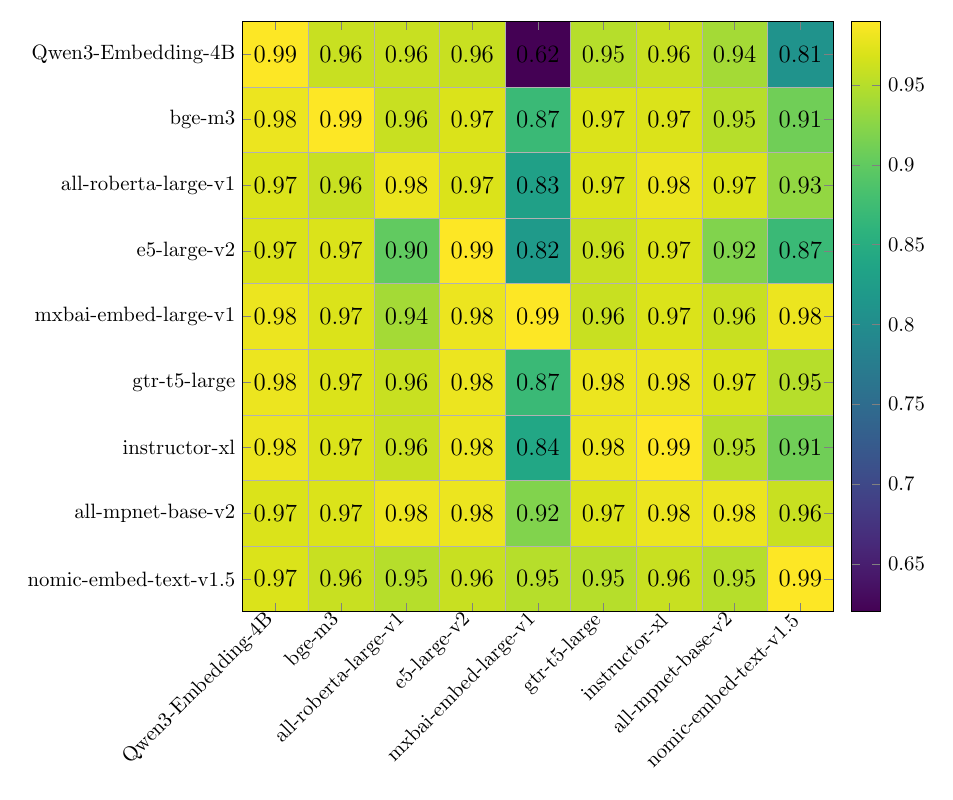}
			\caption*{Linear (SGD)}
		\end{subfigure}%
	\end{subfigure}
	\caption{Task-transfer accuracy matrices for AG News and DBpedia (rows: source models; columns: target models).}
	\label{fig:paper-accuracy-cross-model}
\end{figure}

Table~\ref{tab:paper-overall-performance} and Figure~\ref{fig:paper-accuracy-cross-model} together reveal both the aggregate patterns and the pairwise structure of cross-model transfer. Three observations stand out. Table~\ref{tab:paper-overall-performance} shows both the average of cross-model accuracy (9x9) and the average of the same-model accuracy (the 9 models where the translation is $T_{i\to i}$). Note that this is different from not applying the translation at all; even for the same model we applied the lossy translation (That is why the results are different according to the translation method, in the table).

\emph{Linear (SGD)} achieves the highest cross-model accuracy, which is expected given that it is not dependent on selecting the most optimal anchors. Among anchor-based methods, IRP consistently outperforms \emph{Linear (pinv)} and RR, despite the methods being really close in their logic. This suggests that the anchor pruning step presented in IRP model is effective which could have been applied to the 2 other methods as well, but we tried to apply the methods as presented in their original papers, with the only difference in selecting the initial anchor set. This confirms that the choice of anchors can be improved, however as we also trained \emph{Linear (SGD)} with a large paired corpus, we see approximately what other methods can achieve.
\begin{table}[ht]
	\caption{Average accuracy and macro-F1: cross-model pairs / same-model $T_{i \rightarrow i}$. \emph{Linear (SGD)} uses more paired training data than anchor-based methods and serves as a linear approximate upper bound, not a direct comparison.}
	\label{tab:paper-overall-performance}
	\centering
	\resizebox{\textwidth}{!}{
		
\begin{tabular}{lcccccccc}
\toprule
 & \multicolumn{2}{c}{Linear-alignment} & \multicolumn{2}{c}{RelativeRepresentations} & \multicolumn{2}{c}{IRP} & \multicolumn{2}{c}{Learned-Linear} \\
\cmidrule(lr){2-3} \cmidrule(lr){4-5} \cmidrule(lr){6-7} \cmidrule(lr){8-9}
Task & Accuracy & F1 & Accuracy & F1 & Accuracy & F1 & Accuracy & F1 \\
\midrule
sst2 & 70.16/88.75 & 66.39/88.56 & 66.73/\textbf{91.73} & 60.29/\textbf{91.67} & 74.64/91.09 & 72.05/91.04 & \textbf{84.97}/90.92 & \textbf{84.49}/90.83 \\
AG News & 52.45/85.15 & 46.82/84.60 & 33.37/77.58 & 23.97/76.51 & 57.95/87.34 & 53.65/86.91 & \textbf{83.48}/\textbf{89.63} & \textbf{83.06}/\textbf{89.63} \\
emotion & 28.52/57.21 & 24.22/56.10 & 24.28/52.54 & 16.02/50.34 & 32.36/58.19 & 29.35/58.32 & \textbf{43.74}/\textbf{59.72} & \textbf{42.27}/\textbf{58.67} \\
cola & 63.69/73.93 & 59.12/72.16 & 55.33/57.69 & 41.46/54.13 & 64.07/75.08 & 60.38/72.16 & \textbf{68.04}/\textbf{75.71} & \textbf{64.69}/\textbf{75.04} \\
dbpedia & 44.02/88.19 & 40.88/87.24 & 17.62/73.51 & 12.84/72.36 & 50.90/94.59 & 48.16/94.25 & \textbf{89.11}/\textbf{96.97} & \textbf{88.72}/\textbf{96.89} \\
imdb & 60.46/80.12 & 52.66/79.16 & 53.89/75.90 & 43.83/74.11 & 63.65/\textbf{81.39} & 58.36/\textbf{81.02} & \textbf{73.72}/80.67 & \textbf{71.67}/79.87 \\
mrpc & 51.74/57.04 & 43.76/54.36 & 52.62/56.24 & 39.89/52.59 & 51.93/57.36 & 45.63/54.33 & \textbf{55.08}/\textbf{58.09} & \textbf{50.27}/\textbf{55.72} \\
qqp & 52.88/69.69 & 42.87/67.54 & 53.82/69.86 & 43.56/67.23 & 52.55/64.07 & 44.81/61.24 & \textbf{54.47}/\textbf{72.55} & \textbf{46.85}/\textbf{70.26} \\
\bottomrule
\end{tabular}

	}
\end{table}

The pairwise matrices in Figure~\ref{fig:paper-accuracy-cross-model} show that performance is not uniform. Compatible clusters align with shared architectural and training choices: \texttt{gtr-t5-large}–\texttt{instructor-xl} transfer well bidirectionally due to matching T5 backbones. \texttt{e5-large-v2} is consistently a poor source and target, but not necessarily always the least compatible, as for some tasks/some translation methods, there are worse performing pairs. \texttt{Qwen3-Embedding-4B}, despite its causal decoder backbone, which might be the most different from the rest, still shows moderate compatibility. This suggests that architectural difference alone is not a sufficient condition for compatibility.

Overall, the cross-model performance (outside the diagonal) is often lower than the same-model performance (diagonal), and sometimes much lower for certain pairs/tasks. This questions the universality hypothesis, and the suggestion that the representations of different models, regardless of their architecture, training objective, pooling strategy, etc., are similar up to a simple transformation (e.g., linear).

\subsection{Representation Fidelity After Translation}

\begin{figure}[ht]
	\centering
	\begin{subfigure}[b]{\textwidth}
		\caption{Linear (pinv)}
		\centering
		\resizebox{0.9\textwidth}{!}{
			\begin{subfigure}{0.57\textwidth}
				\adjincludegraphics[width=\textwidth, trim={0 0 {0.18\width} {0.07\height}}, clip]{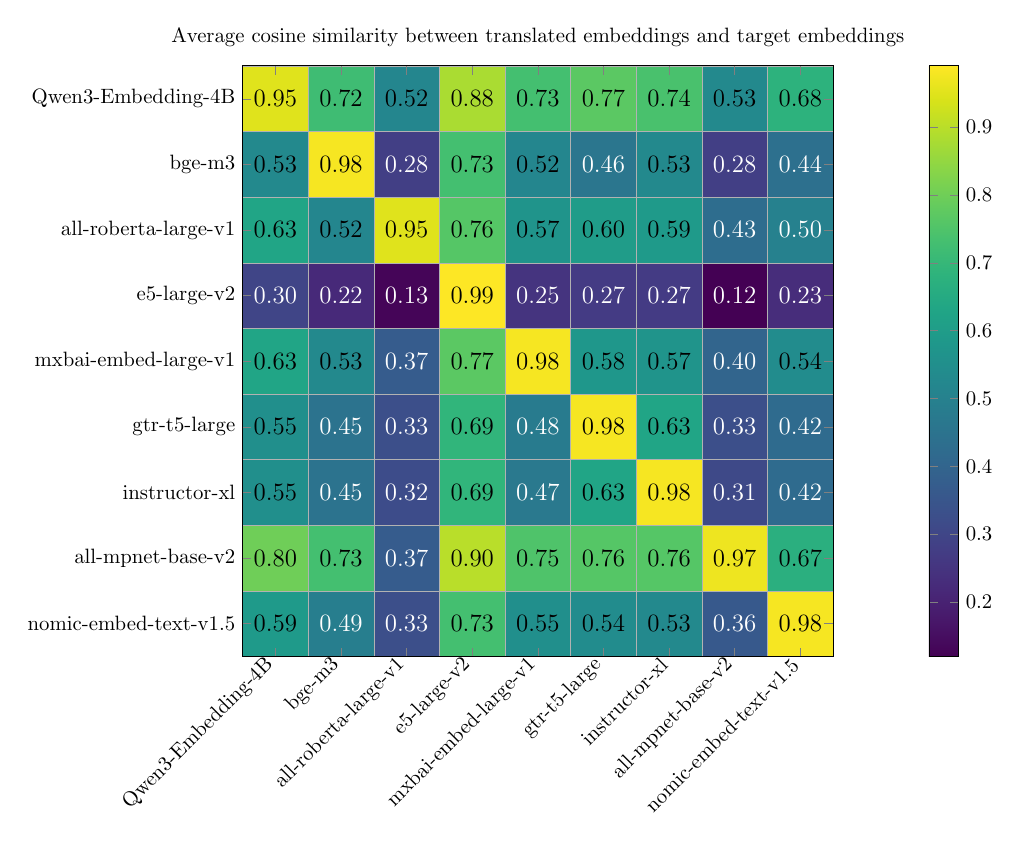}
				\caption*{Cosine similarity}
			\end{subfigure}%
			\hspace{2em}
			\begin{subfigure}{0.425\textwidth}
				\adjincludegraphics[width=\textwidth, trim={{0.22\width} 0 {0.2\width} {0.07\height}}, clip]{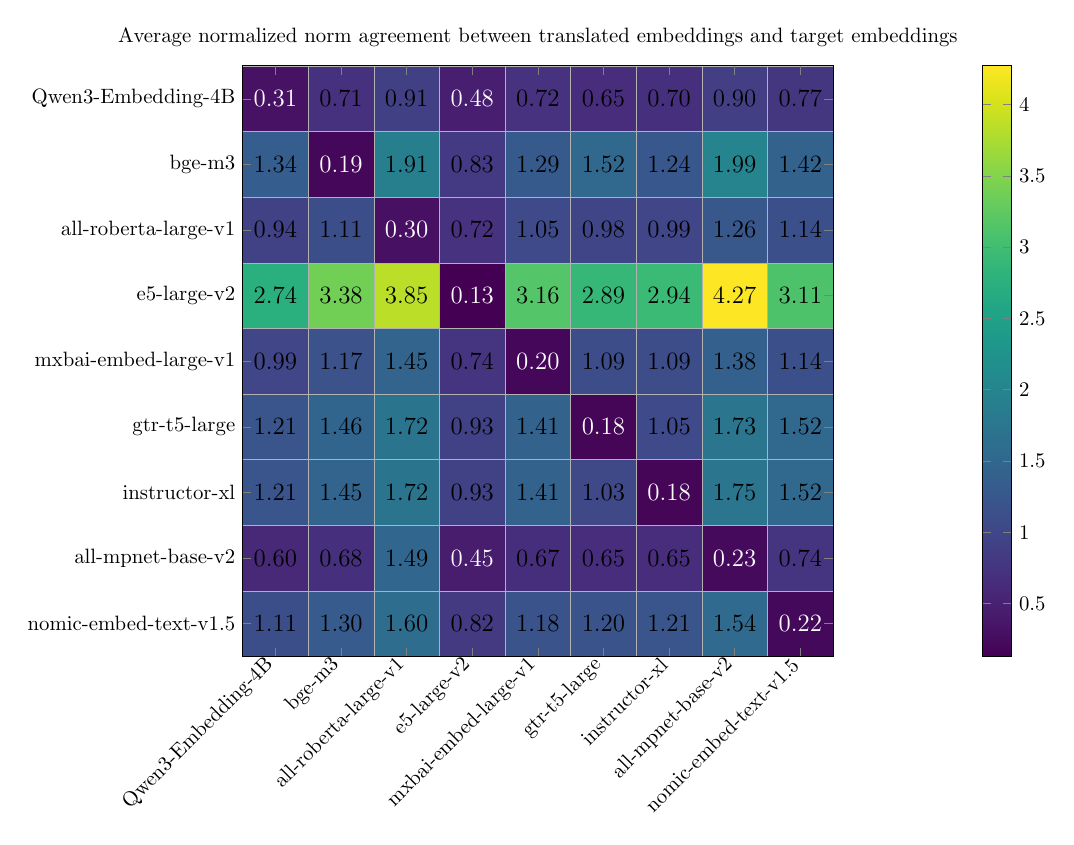}
				\caption*{Normalized error norm}
			\end{subfigure}
		}
	\end{subfigure}
	\begin{subfigure}[b]{\textwidth}
		\caption{Linear (SGD)}
		\centering
		\resizebox{0.9\textwidth}{!}{
			\begin{subfigure}{0.57\textwidth}
				\adjincludegraphics[width=\textwidth, trim={0 0 {0.18\width} {0.07\height}}, clip]{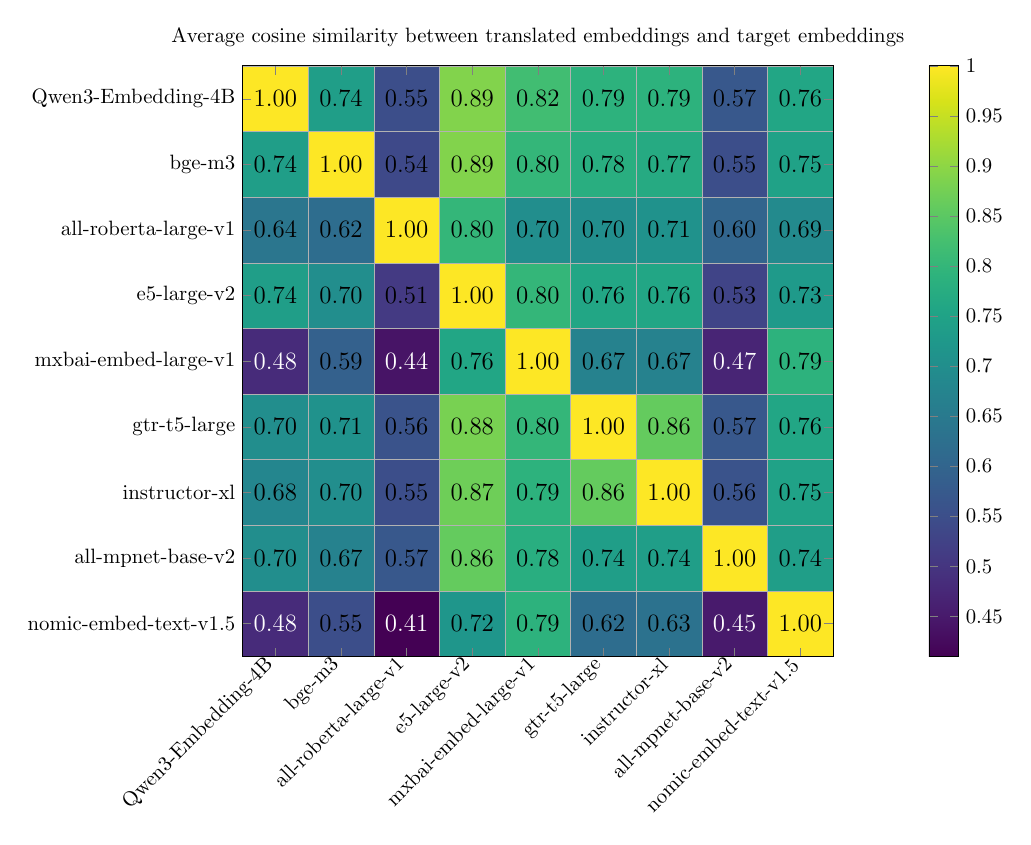}
			\end{subfigure}%
			\hspace{2em}
			\begin{subfigure}{0.425\textwidth}
				\adjincludegraphics[width=\textwidth, trim={{0.22\width} 0 {0.2\width} {0.07\height}}, clip]{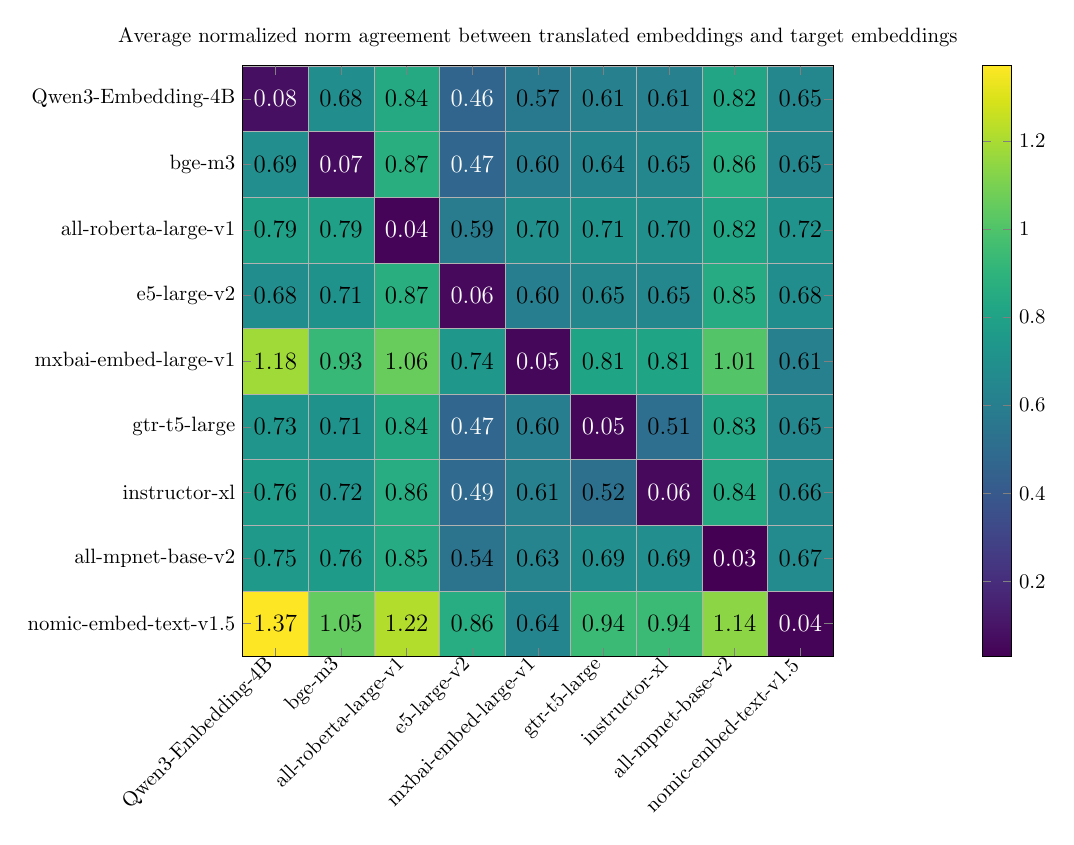}
			\end{subfigure}
		}
	\end{subfigure}
	\caption{Embedding fidelity on held-out words. T5-family pairs show the highest cosine similarity.}
	\label{fig:paper-similarity-matrices}
\end{figure}

We compare translated and original target embeddings for 1000 common English words held out from the anchor set, reporting cosine similarity and normalized error norm for \emph{Linear (pinv)} and \emph{Linear (SGD)}. Figure~\ref{fig:paper-similarity-matrices} presents the mean similarity/distance across the 1000 words for each model pair.

The fidelity matrices largely mirror the CKA and downstream transfer patterns. High cosine similarity and low normalized error appear for the T5-family pair; low fidelity appears for pairs involving \texttt{e5-large-v2} (note that the cosine similarity target in \texttt{e5-large-v2} column is [0.7, 1], so it might appear higher than the others, but it is not, and some translations are even out of that range, making the translation further different from the original).

The correspondence between fidelity and downstream transfer exists but does not explain everything. Indeed, evaluating fidelity might be more important, as it shows the imperfection even more. As for some of the evaluated tasks, they might not need the exact translation and can handle the small error margin, but for other tasks, this error might have a bigger impact. For example for high prompt adherent image generation tasks, or text-to-text generation tasks that need an exact answer not an approximation. Here, even with the linear map trained via SGD, the translation is far from perfect. This further calls the universality hypothesis into question.

\subsection{$k$-NN Retrieval}

We evaluate top-$k$ retrieval ($k \in \{1,5,10\}$) on 1000 common English words using Basic English word anchors (Figure~\ref{fig:paper-knn-words}); sentence-anchor results for Wikitext-103 are provided in Appendix~\ref{app:knn-sentences}.

\begin{figure}[ht]
	\centering
	\begin{subfigure}{\textwidth}
		\caption{RR}
		\begin{subfigure}[t]{0.391\textwidth}
			\adjincludegraphics[width=\textwidth, trim={0 0 {0.1\width} 0}, clip]{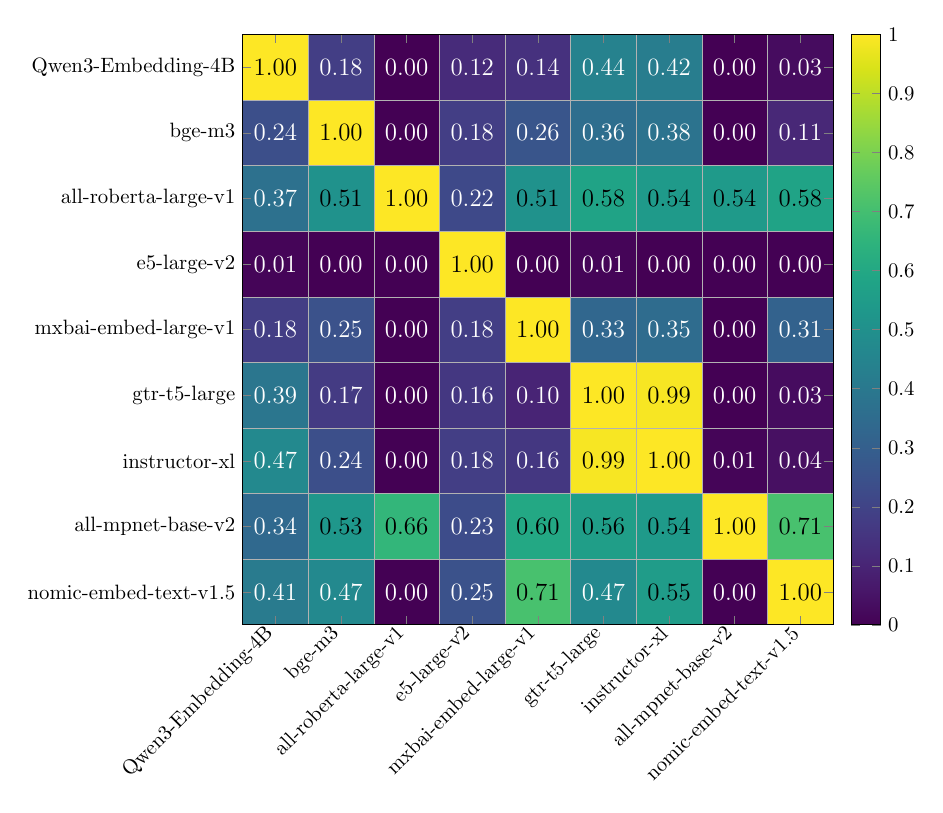}
			\caption*{Top-1}
		\end{subfigure}%
		\begin{subfigure}[t]{0.28\textwidth}
			\adjincludegraphics[width=\textwidth, trim={{0.255\width} 0 {0.1\width} 0}, clip]{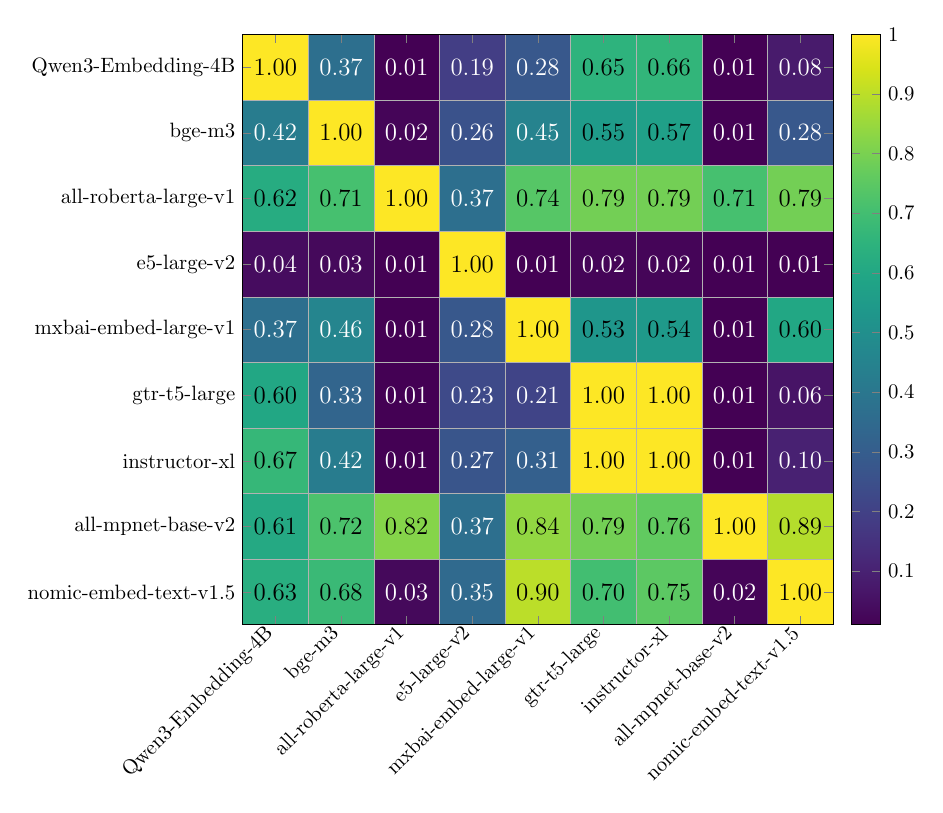}
			\caption*{Top-5}
		\end{subfigure}%
		\begin{subfigure}[t]{0.3235\textwidth}
			\adjincludegraphics[width=\textwidth, trim={{0.255\width} 0 0 0}, clip]{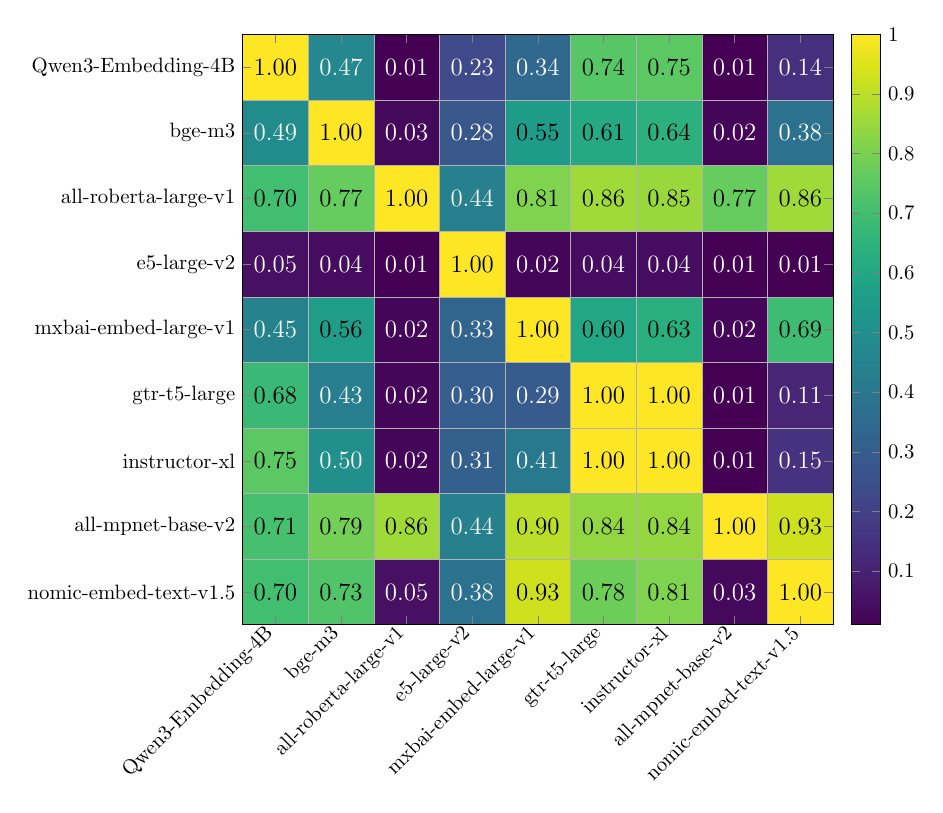}
			\caption*{Top-10}
		\end{subfigure}
	\end{subfigure}
	\hfill
	\begin{subfigure}{\textwidth}
		\caption{Linear (pinv)}
		\begin{subfigure}[t]{0.391\textwidth}
			\adjincludegraphics[width=\textwidth, trim={0 0 {0.1\width} 0}, clip]{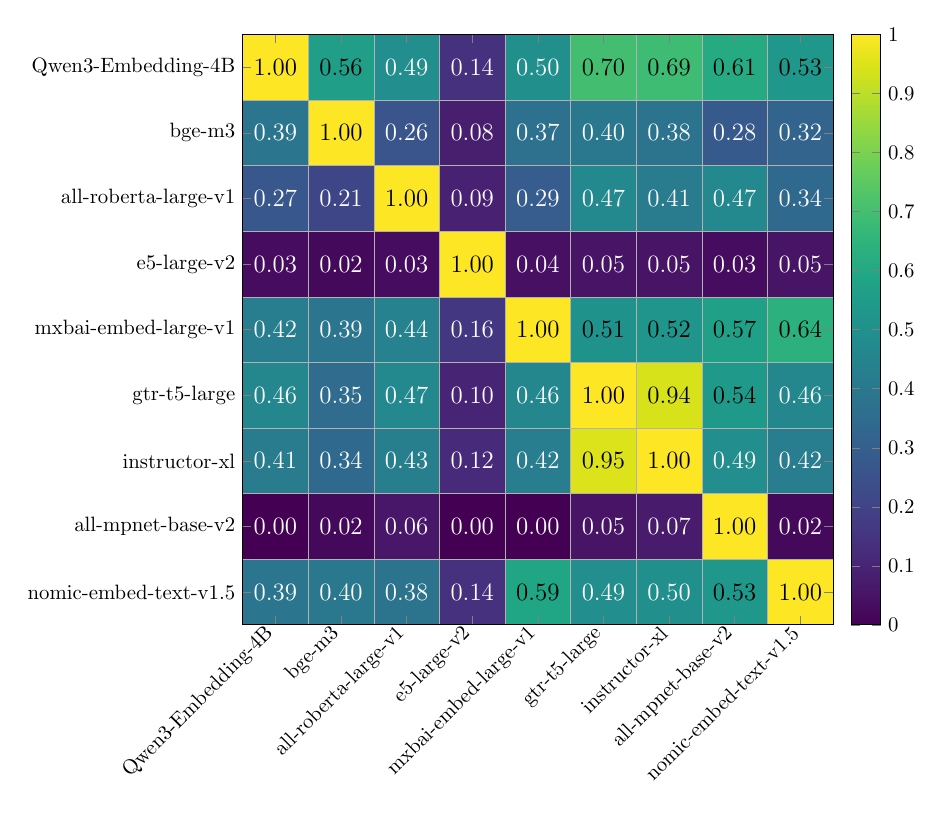}
			\caption*{Top-1}
		\end{subfigure}%
		\begin{subfigure}[t]{0.28\textwidth}
			\adjincludegraphics[width=\textwidth, trim={{0.255\width} 0 {0.1\width} 0}, clip]{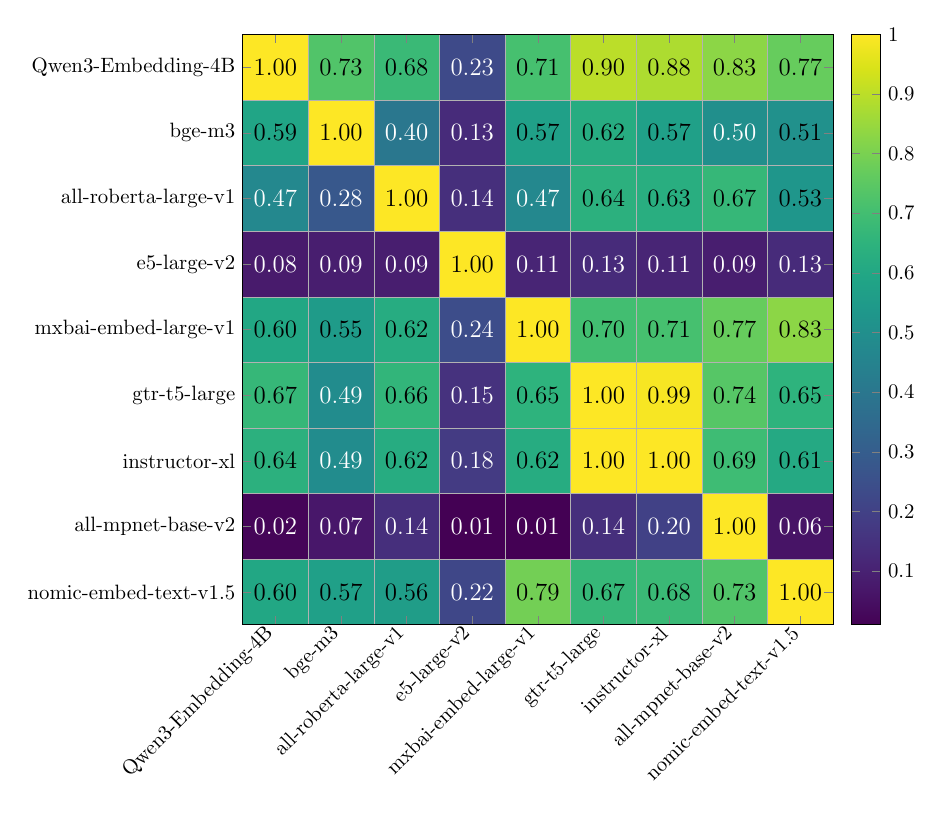}
			\caption*{Top-5}
		\end{subfigure}%
		\begin{subfigure}[t]{0.3235\textwidth}
			\adjincludegraphics[width=\textwidth, trim={{0.255\width} 0 0 0}, clip]{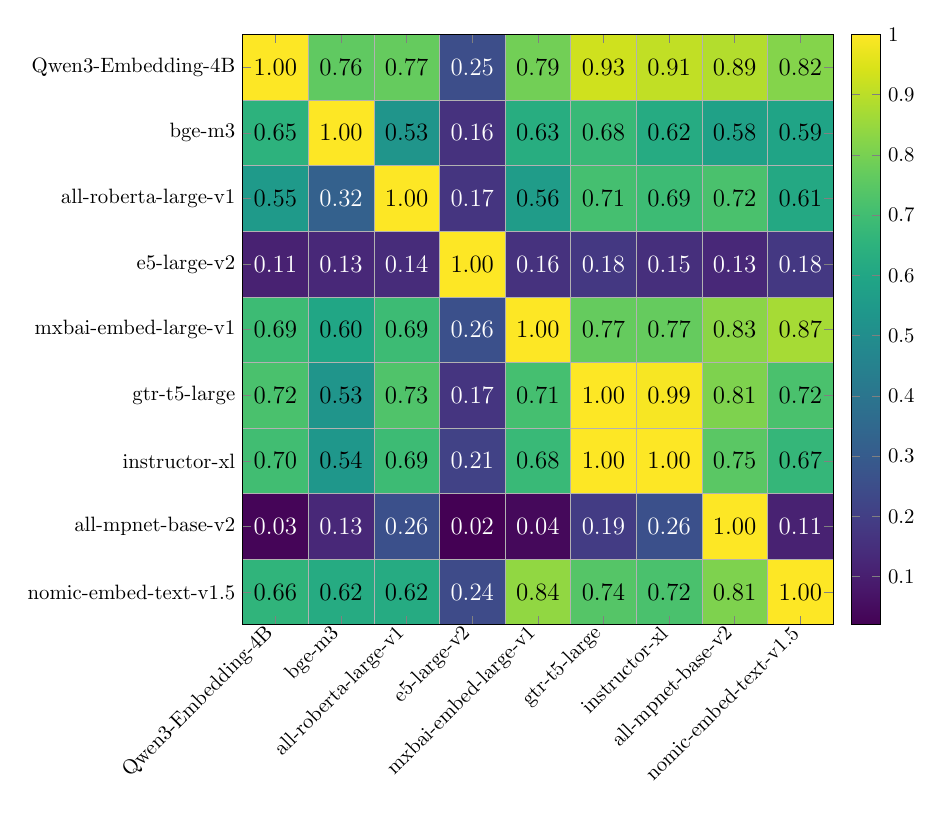}
			\caption*{Top-10}
		\end{subfigure}
	\end{subfigure}
	\caption{$k$-NN retrieval on 1000 common words (Basic English anchors). The compatibility clusters visible in downstream transfer, T5-family, BERT mean-pooling, recur here. \texttt{e5-large-v2} and \texttt{mxbai-embed-large-v1} show degraded retrieval as both source and target.}
	\label{fig:paper-knn-words}
\end{figure}

The retrieval matrices echo the downstream transfer pattern: T5-family retrieve well; \texttt{e5-large-v2} remain outliers. Retrieval and transfer do not always agree: k-NN measures average neighborhood geometry while classification depends on preserving decision boundaries. Overall, retrieval with $k$-NN shows similar patterns to other downstream tasks in terms of model compatibility. However, $k$-NN retrieval might not be the best way to evaluate translation quality as it might suffer from the hubness problem, where some points (hubs) appear as nearest neighbors very frequently to other points \cite{radovanovi&263HubsSpacePopular2010} (refer to Appendix~\ref{app:hubness} for deep hubness analysis in the embeddings of the tested models). Nevertheless, it is still worth evaluating to understand which models are compatible and which architectural or training differences make them incompatible.

\section{Model Compatibility}

To summarize pairwise behavior, we define a symmetric compatibility score $\mathrm{Compat}(A,B) = \tfrac{1}{2}(\mathrm{Acc}_{A\rightarrow B} + \mathrm{Acc}_{B\rightarrow A})$, the average accuracy in both translation directions across all methods and tasks between models $A$ and $B$.

\begin{figure}[ht]
	\centering
	\resizebox{0.9\textwidth}{!}{
		\includegraphics[width=\textwidth]{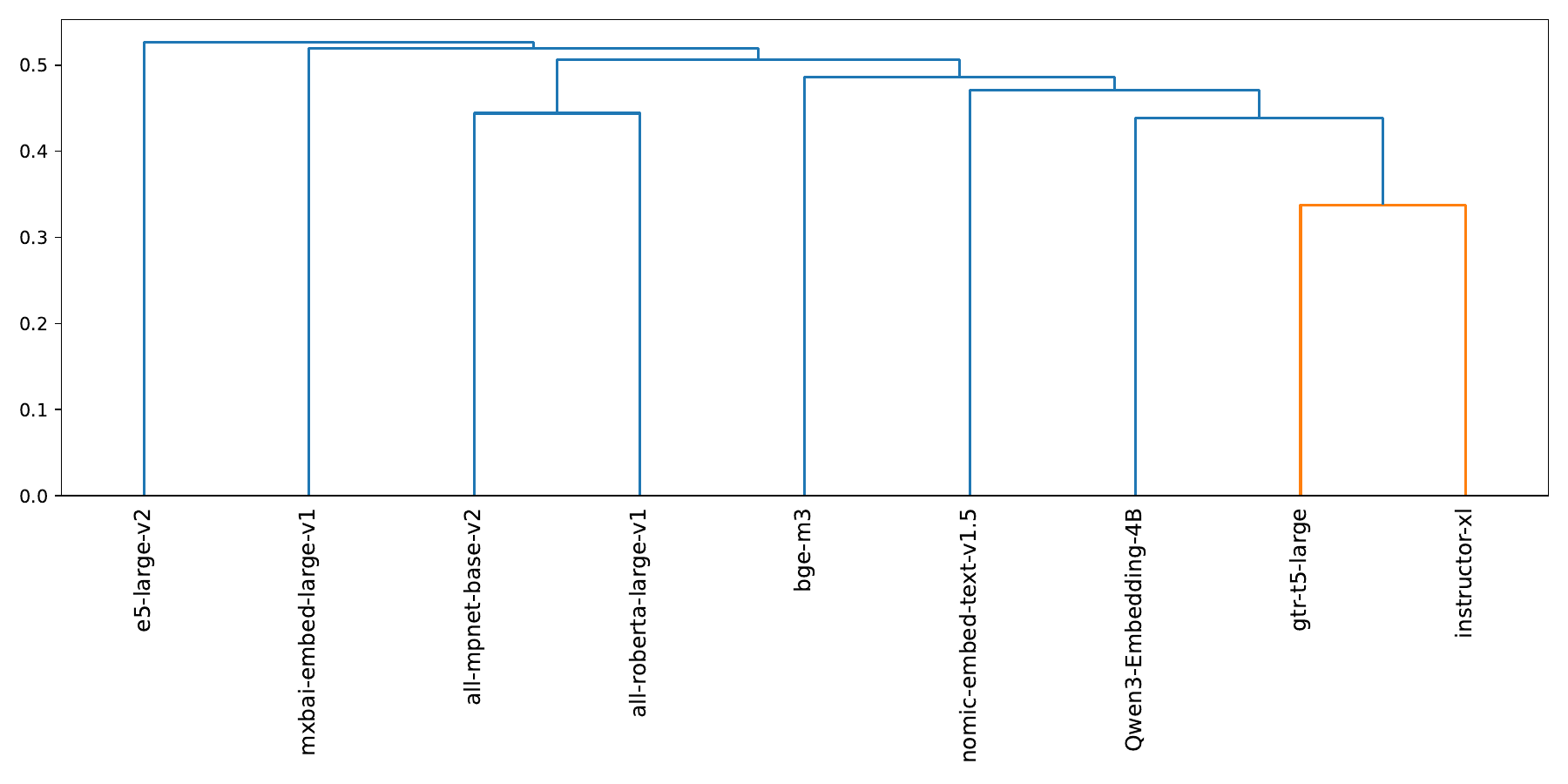}
	}
	\caption{Compatibility dendrogram from bidirectional IRP transfer accuracy.}
	\label{fig:paper-compatibility-dendrogram}
\end{figure}

Figure~\ref{fig:paper-compatibility-dendrogram} crystallizes the compatibility structure. The two tightest clusters, \texttt{gtr-t5-large}/\texttt{instructor-xl} and \texttt{all-mpnet-base-v2}/\texttt{all-roberta-large-v1}, are explained by shared inductive biases: same encoder family, same pooling (mean), and similar contrastive training. Models sharing two or more of these properties cluster together; those differing on training objective, pooling, or normalization cluster farther apart.

Two exceptions are informative. \texttt{Qwen3-Embedding-4B} (causal decoder, last-token pooling, 2560-dim) clusters moderately close to the encoder group, closer than \texttt{e5-large-v2}, a standard BERT model. Its contrastive fine-tuning produces a semantically organized space broadly compatible with encoder-style models: training objective overrides architectural differences. \texttt{e5-large-v2}, conversely, shares architecture, pooling, and normalization with \texttt{all-roberta-large-v1} yet is the most isolated node, due to its E5 pretraining specificities \cite{wangTextEmbeddingsWeaklySupervised2024}. Compatibility is shaped by the combination of architecture, training objective, data distribution, and pooling strategy.

The fact that some models are more compatible with each other than with others, and that this compatibility varies with the task and translation method, further emphasizes that the representations of different models are not as similar as suggested.

\section{Discussion}

Simple transformations explain a meaningful part of the relationship between text embedding spaces: they recover shared local structure, reveal compatibility clusters, and enable non-trivial downstream transfer. The analysis identifies three levels at which compatibility breaks down. At the \emph{geometric} level, models trained on different data distributions or with different pooling strategies produce spaces that can be related by a single linear map, but only to a certain extent. At the \emph{task} level, even geometrically similar models can fail on fine-grained tasks.

Furthermore, the relatively good cross-model performance of \emph{Linear (SGD)} on certain tasks confirms that the universality hypothesis is not completely untrue. Maybe sticking to very simple translation methods of the linear family is too much of a limitation, and allowing even small non-linearities would allow bridging the gap between more models, and lend more support to the universality hypothesis.

\section{Limitations and Broader Impact}
\label{sec:limitations}

Our work has the following limitations: (1) we evaluate only simple transformation families over a finite model set with classification-oriented tasks; (2) the fixed SVM setup keeps comparisons controlled but may understate absolute performance. No uncertainty estimates or compute accounting are reported.

A more fundamental limitation is the scope of the translation problem itself. All methods evaluated here operate on fixed-length sentence or word embeddings: the map is $\mathbb{R}^{d_A} \to \mathbb{R}^{d_B}$, a single-vector-to-single-vector problem. Many practically important tasks, cross-model attention, token-level generation, sequence-to-sequence reuse, require translating full contextual representations of shape $\mathbb{R}^{l_A \times d_A} \to \mathbb{R}^{l_B \times d_B}$, where $l_A$ and $l_B$ are sequence lengths. When the two models use different tokenizers, $l_A \neq l_B$ in general for the same input, creating a structural misalignment that a position-wise linear map cannot resolve. Even with the same tokenizer, cross-attention-based decoders or self-attention-based text-to-text decoders depend on the full key-value sequence from the encoder rather than on a pooled summary; the geometric structure exploited by our translators simply does not apply in such cases. These sequence-level challenges represent a qualitatively harder translation problem that linear methods are ill-equipped to address, and that future work will need to confront directly.

\section{Conclusion}

The main conclusion of this work is that simple transformations work, but only in some cases, and the failures are as telling as the successes. Across our nine-model evaluation, cross-model transfer achieves competitive performance for a handful of compatible pairs, those sharing architecture family, training objective, pooling strategy, and data distribution. However, for many pairs—especially those crossing architectural or training boundaries—linear translation degrades substantially. The downstream accuracy and macro-F1 averages are often markedly worse than in-model performance, and the confusion matrices tell a sharper story: off-diagonal degradation is not uniform noise but structured failure, entire class boundaries collapse under translation for incompatible pairs, while staying intact for compatible ones.

These results push back against strong universality hypotheses in the literature. Work on relative representations \cite{moschellaRelativeRepresentationsEnable2023b}, latent space translation \cite{maiorcaLatentSpaceTranslation2023a, maiorcaLatentSpaceTranslation2024b}, and the Platonic Representation Hypothesis \cite{huhPlatonicRepresentationHypothesis2024a} has suggested that independently trained models may converge toward shared representational geometry. Our findings qualify this view: in the controlled setting of sentence and word embeddings, with models deliberately chosen to span diverse architectures and training regimes, the geometry is \emph{not} universally shared. The pairs that transfer well are exceptions explained by specific shared design choices, not evidence of a general convergence. Our results show that, beyond simple settings like MNIST or CIFAR classifiers where universality results are strongest, the hypothesis may not extend to heterogeneous, large-scale text embedding models trained on vastly different corpora.

The conclusion for future work is that more complex translators are needed. Linear maps are useful as diagnostic baselines precisely because they make the cost of incompatibility visible; but bridging genuinely heterogeneous spaces will require nonlinear or learned alignment mechanisms that go beyond what a pseudo-inverse over a small word anchor set can provide.

\section*{Acknowledgments}

This research was funded in part by the SNS JU 6GARROW project under the European Union's Horizon programme, Grant Agreement No. 101192194, in collaboration with the Republic of Korea (ROK).

\FloatBarrier
\printbibliography

\appendix

\section{Analogy-Style Reconstruction Across Heterogeneous Encoders}
\label{app:analogy}

This appendix reports the empirical analogy-reconstruction analysis referenced in Section~\ref{sec:local-structure}. For each model we fit coefficients $\alpha, \beta, \gamma$ to minimize $\|\mathrm{king} - (\alpha\,\mathrm{queen} + \beta\,\mathrm{man} + \gamma\,\mathrm{woman})\|_2^2$, and repeat for \textit{queen}. The goal is not to establish identical coefficients but to test whether sign patterns and coarse weightings remain stable across heterogeneous architectures, that is, whether the local linear geometry that Section~\ref{sec:local-structure} relies on actually carries over to modern sentence-level encoders.

\begin{figure}[ht]
	\centering
	\resizebox{0.7\textwidth}{!}{
		\begin{subfigure}[t]{0.575\textwidth}
			\includegraphics[width=\textwidth]{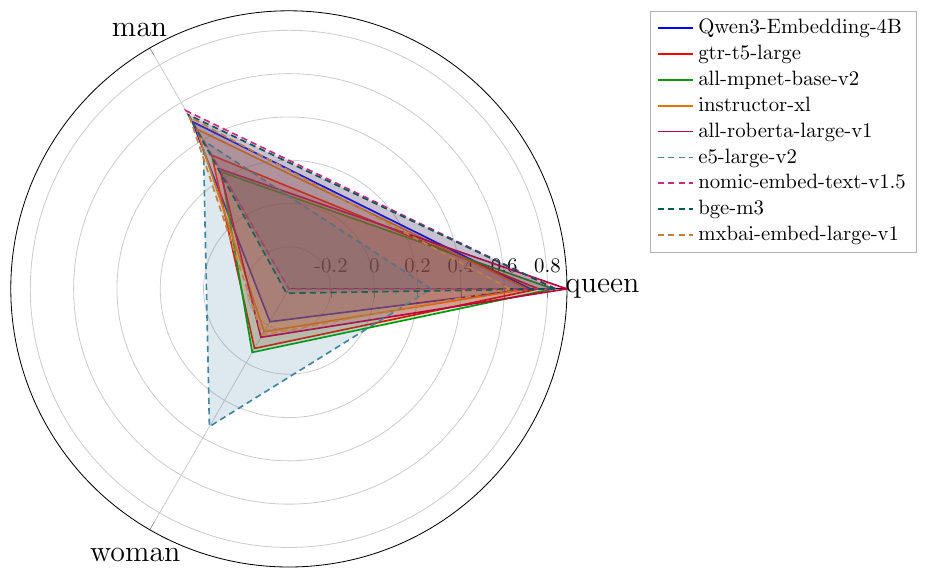}
			\begin{minipage}{0.7\textwidth}
				\captionsetup{justification=raggedright, format=plain, width=\textwidth, singlelinecheck=false}
				\caption{$\mathrm{king} = \alpha\,\mathrm{queen} + \beta\,\mathrm{man} + \gamma\,\mathrm{woman}$.\label{fig:paper-king-q-m-w}}
			\end{minipage}
		\end{subfigure}%
		\hfill
		\begin{subfigure}[t]{0.405\textwidth}
			\adjincludegraphics[width=\textwidth, trim={0 0 {0.3\width} 0}, clip]{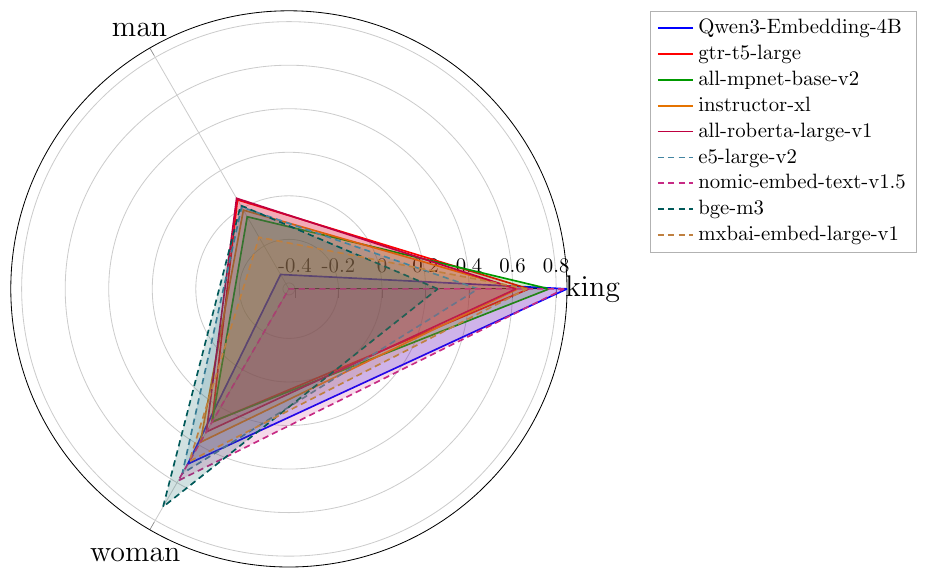}
			\caption{$\mathrm{queen} = \alpha\,\mathrm{king} + \beta\,\mathrm{man} + \gamma\,\mathrm{woman}$.\label{fig:paper-queen-k-m-w}}
		\end{subfigure}
	}
	\caption{Analogy-style decompositions (3-concept basis). Most models agree on sign pattern and coarse weighting.}
	\label{fig:paper-king-queen-3concepts}
\end{figure}

Figure~\ref{fig:paper-king-queen-3concepts} shows broad qualitative agreement: \textit{king} receives positive contributions from \textit{queen} and \textit{man} and a negative one from \textit{woman}; \textit{queen} shows the mirrored pattern. This consistency holds across models, suggesting the royalty-gender subspace is robustly linear across architectures, exactly the prerequisite invoked in Section~\ref{sec:local-structure} for anchor-based translation to be meaningful.

\begin{figure}[ht]
	\centering
	\resizebox{0.7\textwidth}{!}{
		\begin{subfigure}[t]{0.575\textwidth}
			\includegraphics[width=\textwidth]{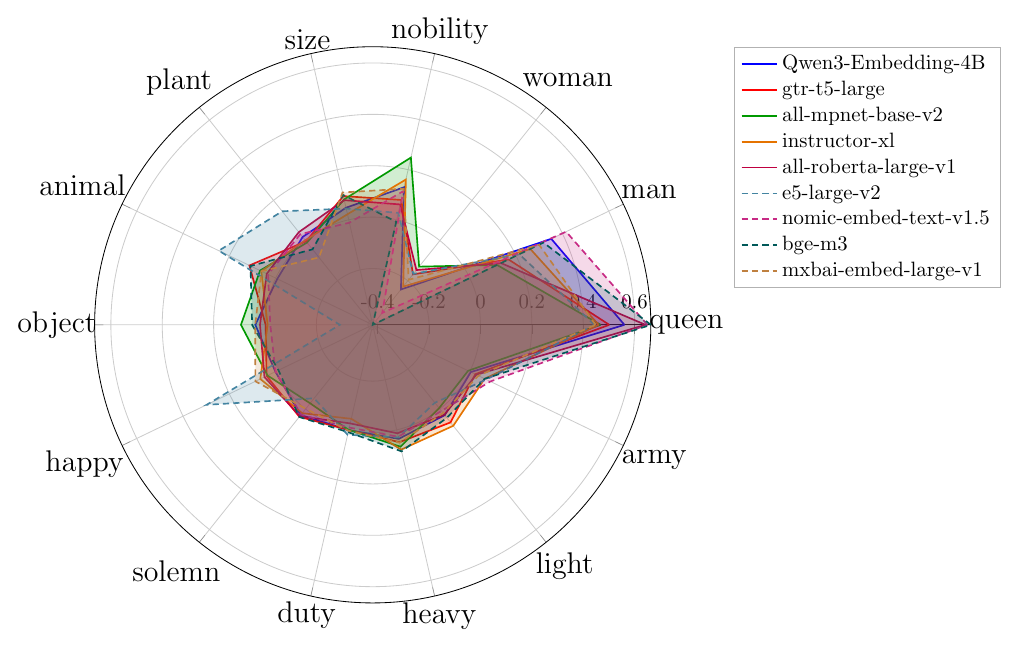}
			\begin{minipage}{0.7\textwidth}
				\captionsetup{justification=raggedright, format=plain, width=\textwidth, singlelinecheck=false}
				\caption{Expanded-basis coefficients for \textit{king}.\label{fig:paper-king-14concepts}}
			\end{minipage}
		\end{subfigure}%
		\hfill
		\begin{subfigure}[t]{0.405\textwidth}
			\adjincludegraphics[width=\textwidth, trim={0 0 {0.3\width} 0}, clip]{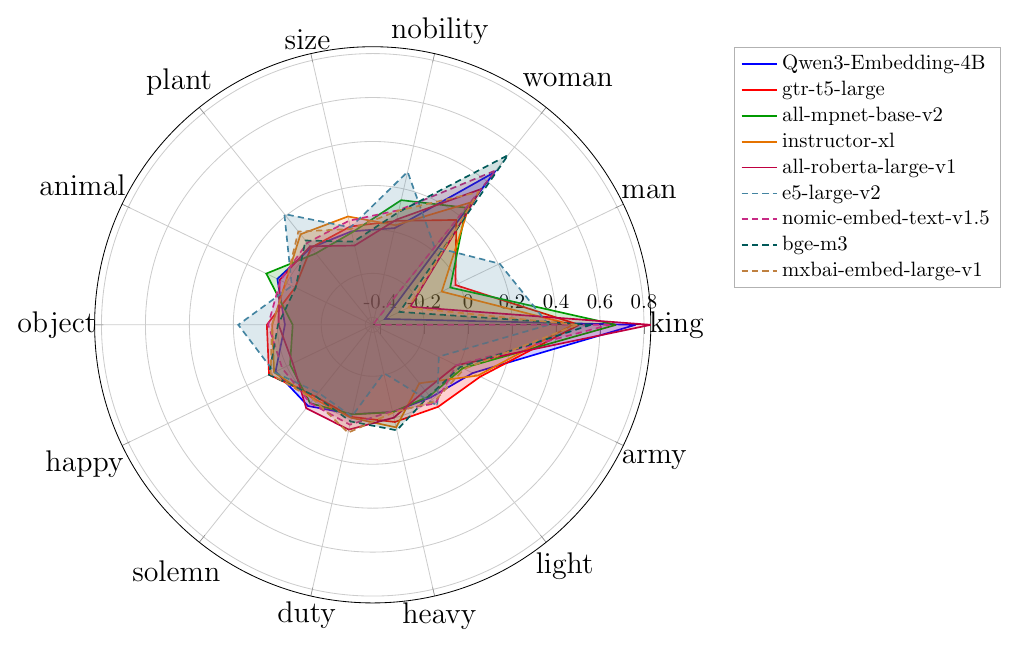}
			\caption{Expanded-basis coefficients for \textit{queen}.\label{fig:paper-queen-14concepts}}
		\end{subfigure}
	}
	\caption{Analogy-style decompositions with a 14-concept basis. Core pattern survives; secondary coefficients become less stable, especially for unnormalized models.}
	\label{fig:paper-king-queen-14concepts}
\end{figure}

Enlarging to a 14-concept basis (Figure~\ref{fig:paper-king-queen-14concepts}) shows the coarse pattern survives, but coefficient-level agreement becomes less stable, particularly for models with different normalization conventions (\texttt{mxbai-embed-large-v1} and \texttt{nomic-embed-text-v1.5} versus the normalized majority). Precise coefficient values are sensitive to each model's scale and normalization, foreshadowing the role these factors play in the cross-model compatibility patterns observed in the main paper.

\section{$k$-NN Retrieval: Wikitext-103 Sentence Anchors}
\label{app:knn-sentences}

\begin{figure}[ht]
	\centering
	\begin{subfigure}[t]{0.391\textwidth}
		\adjincludegraphics[width=\textwidth, trim={0 0 {0.1\width} 0}, clip]{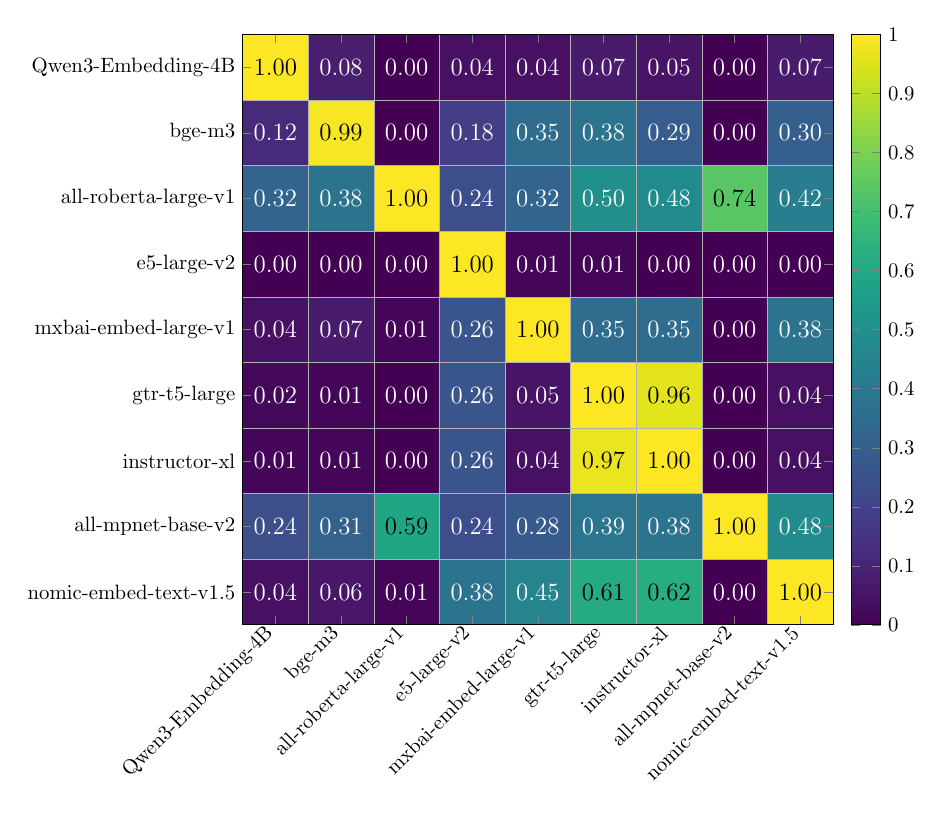}
		\caption*{Top-1}
	\end{subfigure}%
	\begin{subfigure}[t]{0.28\textwidth}
		\adjincludegraphics[width=\textwidth, trim={{0.255\width} 0 {0.1\width} 0}, clip]{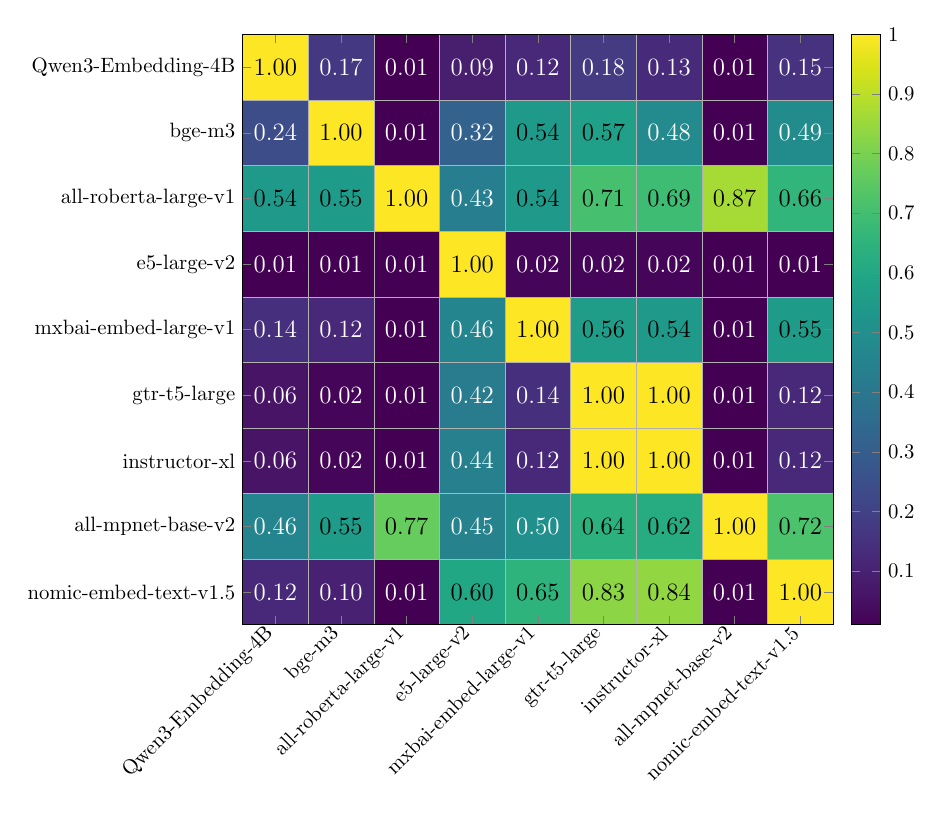}
		\caption*{Top-5}
	\end{subfigure}%
	\begin{subfigure}[t]{0.3235\textwidth}
		\adjincludegraphics[width=\textwidth, trim={{0.255\width} 0 0 0}, clip]{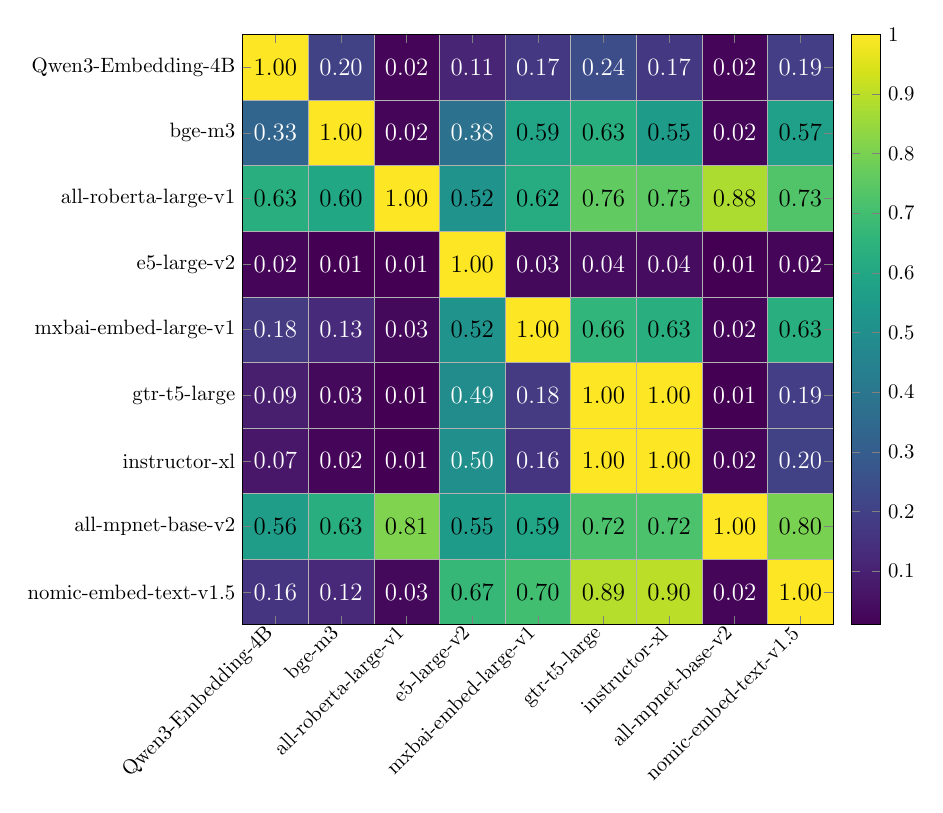}
		\caption*{Top-10}
	\end{subfigure}
	\caption{$k$-NN retrieval: RR, Wikitext-103 sentence anchors.}
	\label{fig:app-knn-sentences-rr}
\end{figure}

\begin{figure}[ht]
	\centering
	\begin{subfigure}[t]{0.391\textwidth}
		\adjincludegraphics[width=\textwidth, trim={0 0 {0.1\width} 0}, clip]{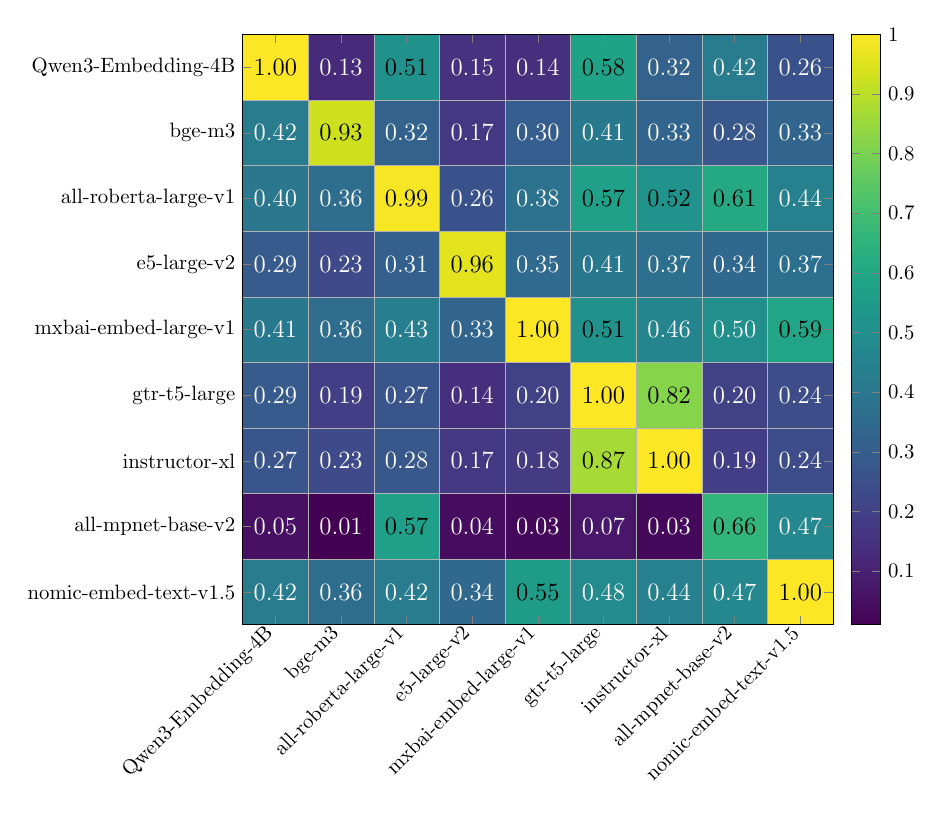}
		\caption*{Top-1}
	\end{subfigure}%
	\begin{subfigure}[t]{0.28\textwidth}
		\adjincludegraphics[width=\textwidth, trim={{0.255\width} 0 {0.1\width} 0}, clip]{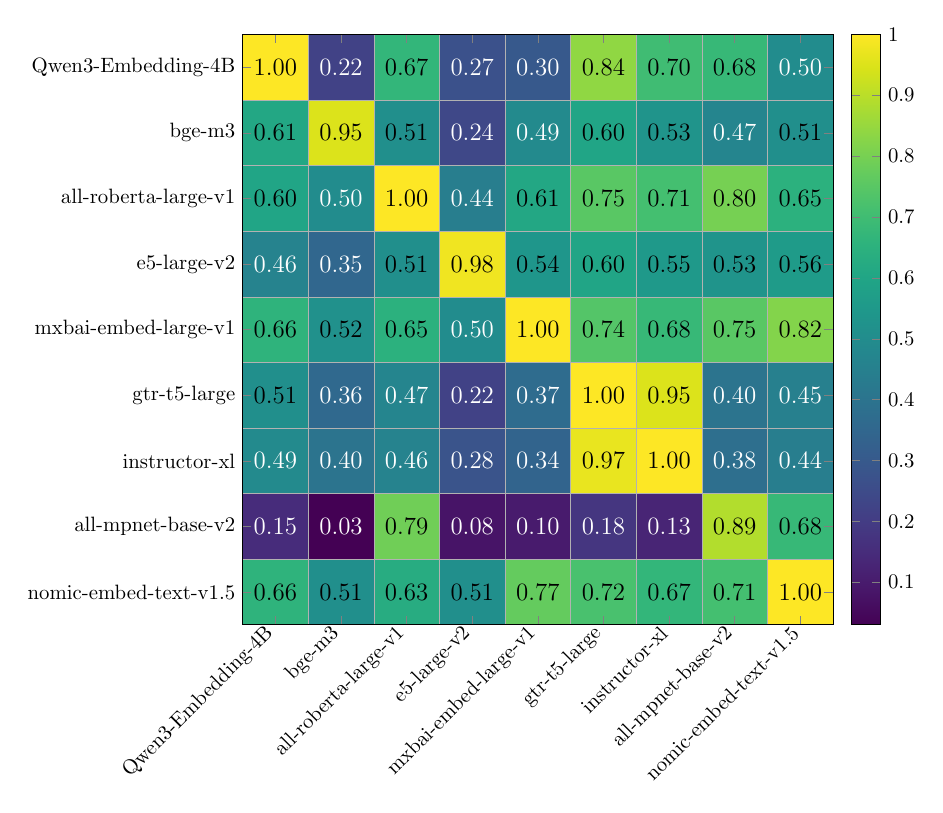}
		\caption*{Top-5}
	\end{subfigure}%
	\begin{subfigure}[t]{0.3235\textwidth}
		\adjincludegraphics[width=\textwidth, trim={{0.255\width} 0 0 0}, clip]{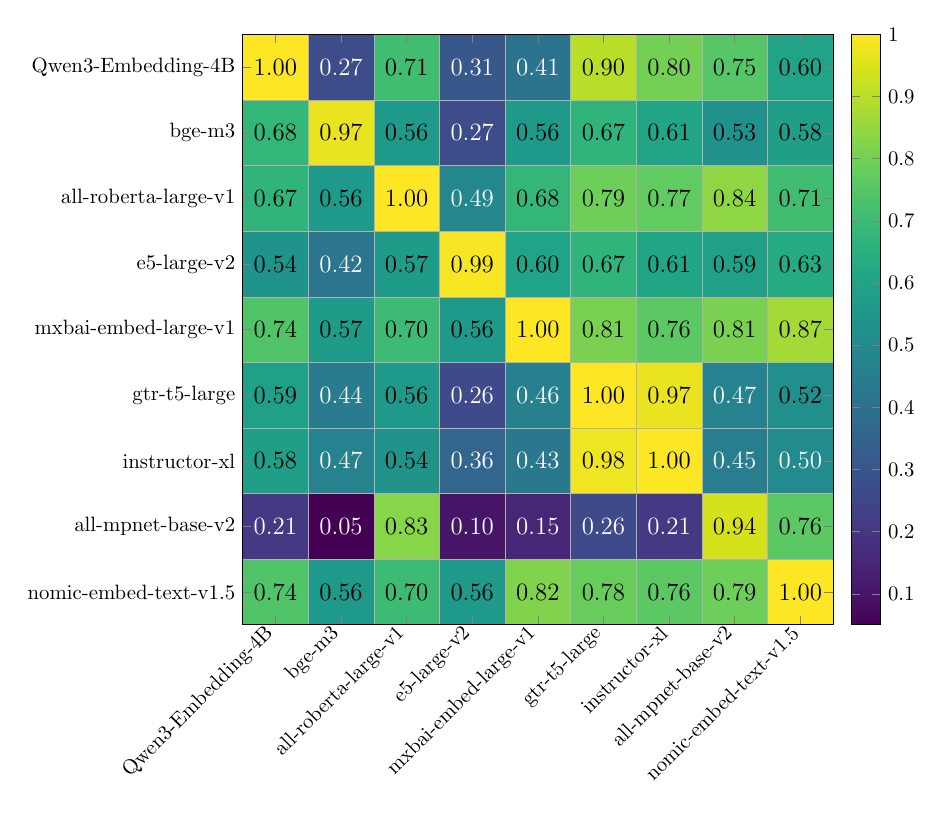}
		\caption*{Top-10}
	\end{subfigure}
	\caption{$k$-NN retrieval: Linear (pinv), Wikitext-103 sentence anchors.}
	\label{fig:app-knn-sentences-pinv}
\end{figure}

\begin{figure}[ht]
	\centering
	\begin{subfigure}[t]{0.391\textwidth}
		\adjincludegraphics[width=\textwidth, trim={0 0 {0.1\width} 0}, clip]{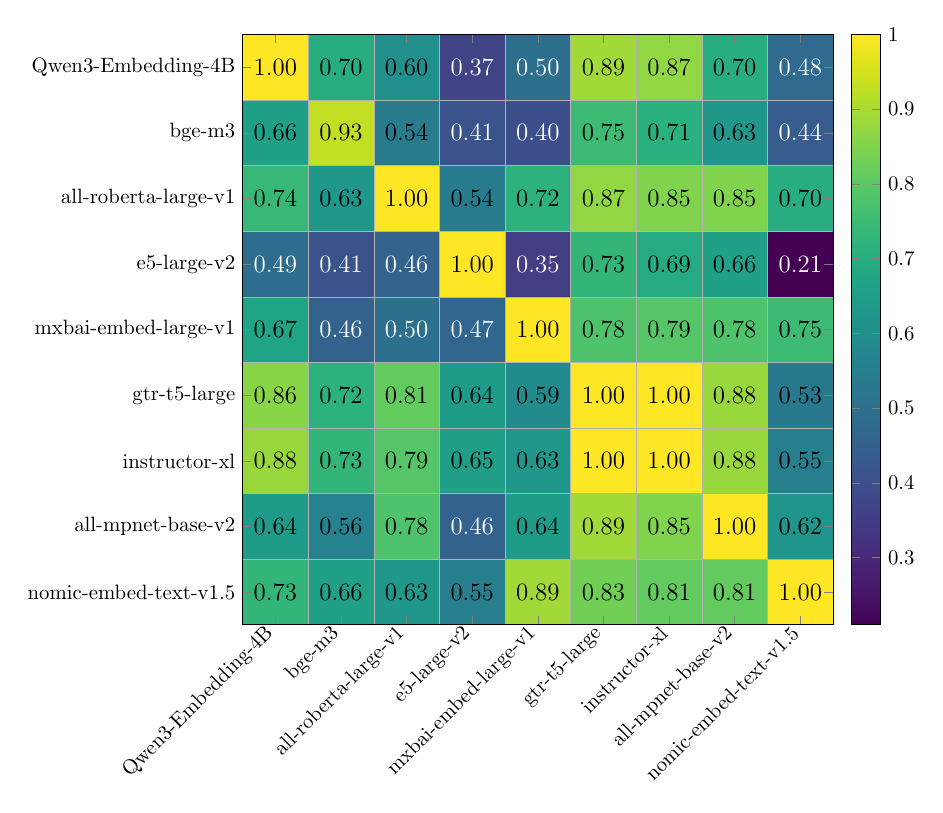}
		\caption*{Top-1}
	\end{subfigure}%
	\begin{subfigure}[t]{0.28\textwidth}
		\adjincludegraphics[width=\textwidth, trim={{0.255\width} 0 {0.1\width} 0}, clip]{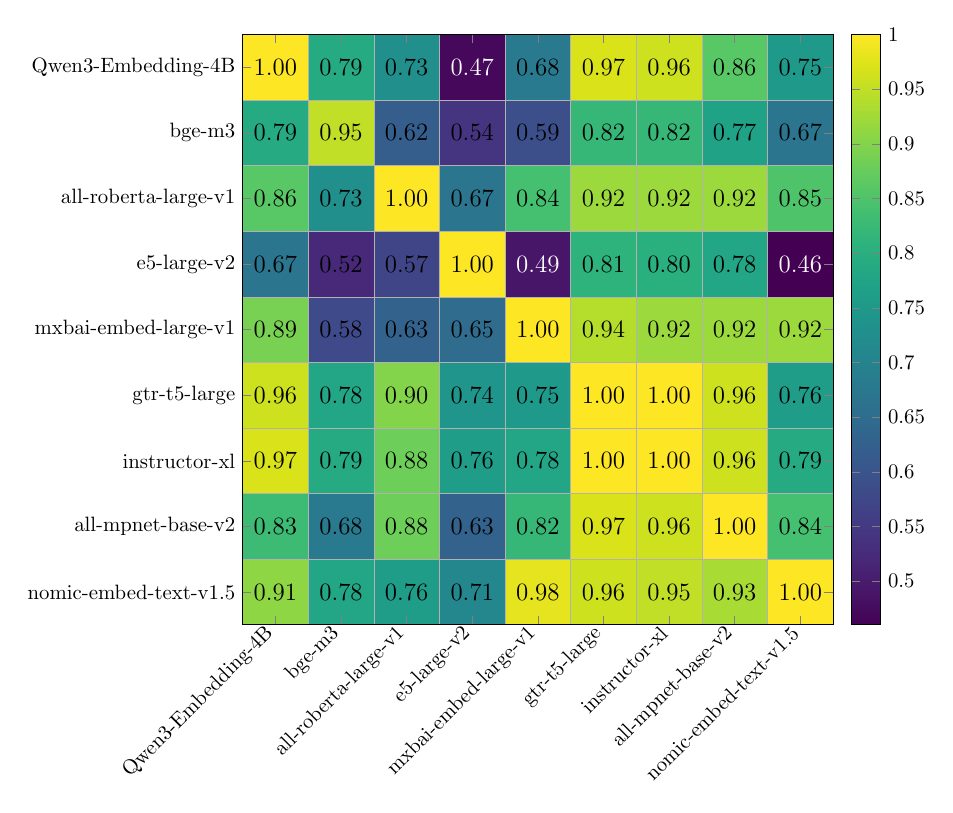}
		\caption*{Top-5}
	\end{subfigure}%
	\begin{subfigure}[t]{0.3235\textwidth}
		\adjincludegraphics[width=\textwidth, trim={{0.255\width} 0 0 0}, clip]{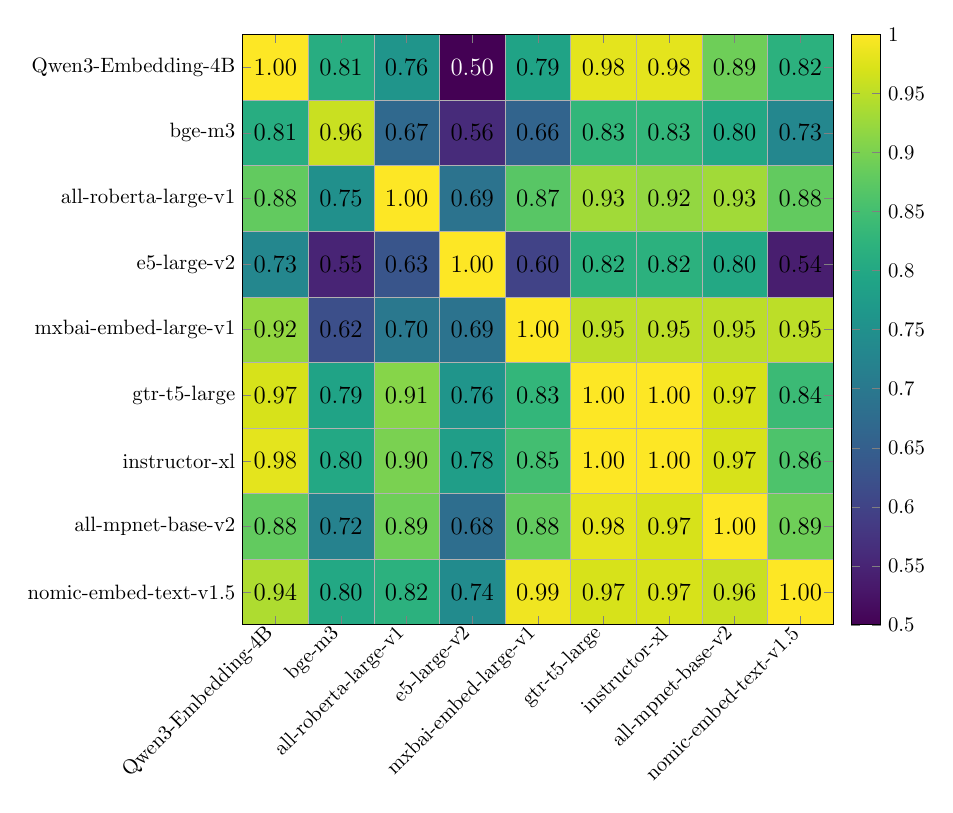}
		\caption*{Top-10}
	\end{subfigure}
	\caption{$k$-NN retrieval: Linear (SGD), Wikitext-103 sentence anchors.}
	\label{fig:app-knn-sentences-sgd}
\end{figure}

We report $k$-NN retrieval results on Wikitext-103 sentence anchors for RR, \emph{Linear (pinv)}, and \emph{Linear (SGD)} in Figures~\ref{fig:app-knn-sentences-rr}--\ref{fig:app-knn-sentences-sgd}. The compatibility structure is consistent with the word-anchor results in the main paper: T5-family retrieve well; \texttt{e5-large-v2} remain outliers. \emph{Linear (SGD)} provides substantially better neighborhood preservation, confirming the upper-bound interpretation, but also suffers in a similar way from model incompatibility.

\section{All Pairwise Accuracy Matrices}
\label{app:accuracy-matrices}
Here we report the full set of task-transfer accuracy matrices for all methods and tasks. The main paper shows a subset of these matrices for AG News and DBpedia, but the patterns are consistent across tasks: compatible pairs show high accuracy; incompatible pairs show structured degradation. Furthermore, for some tasks, the performance of some pairs degrades even more. This suggests that these tasks are more sensitive to the translation fidelity, and that the error margin achieved by the translation methods is not sufficient for these tasks, a finding that again calls the universality hypothesis into question.

\begin{figure}[ht]
	\begin{subfigure}[t]{0.304\textwidth}
		\adjincludegraphics[width=\textwidth, trim={0 0 {0.1\width} 0}, clip]{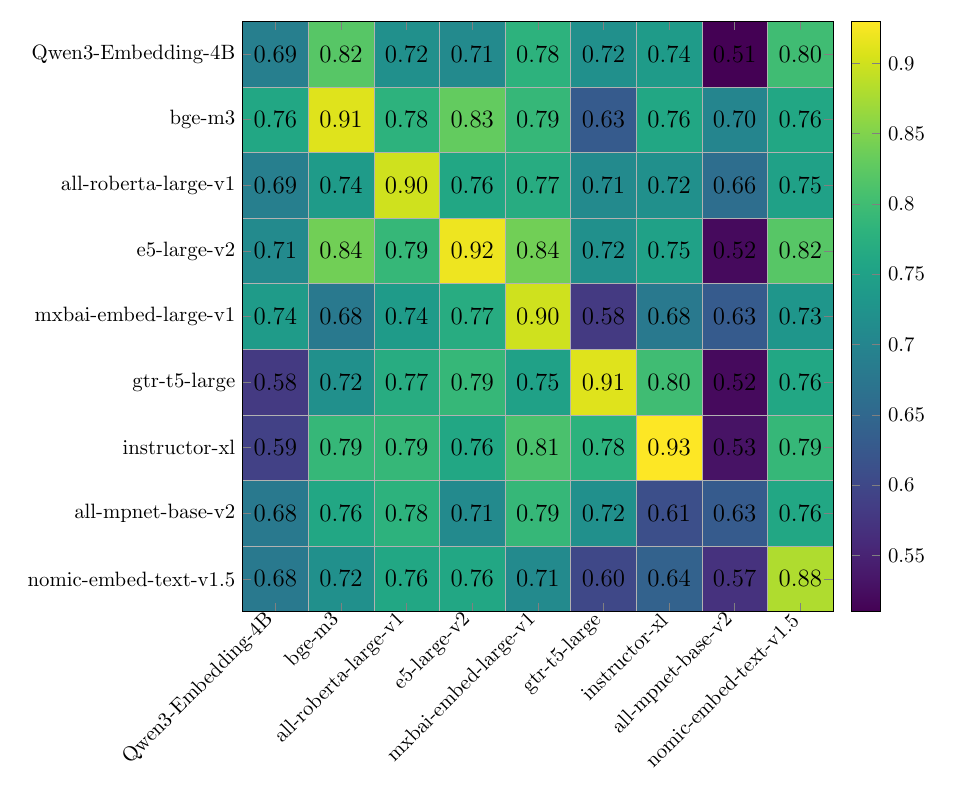}
		\caption*{Linear (pinv)}
	\end{subfigure}%
	\begin{subfigure}[t]{0.22\textwidth}
		\adjincludegraphics[width=\textwidth, trim={{0.25\width} 0 {0.1\width} 0}, clip]{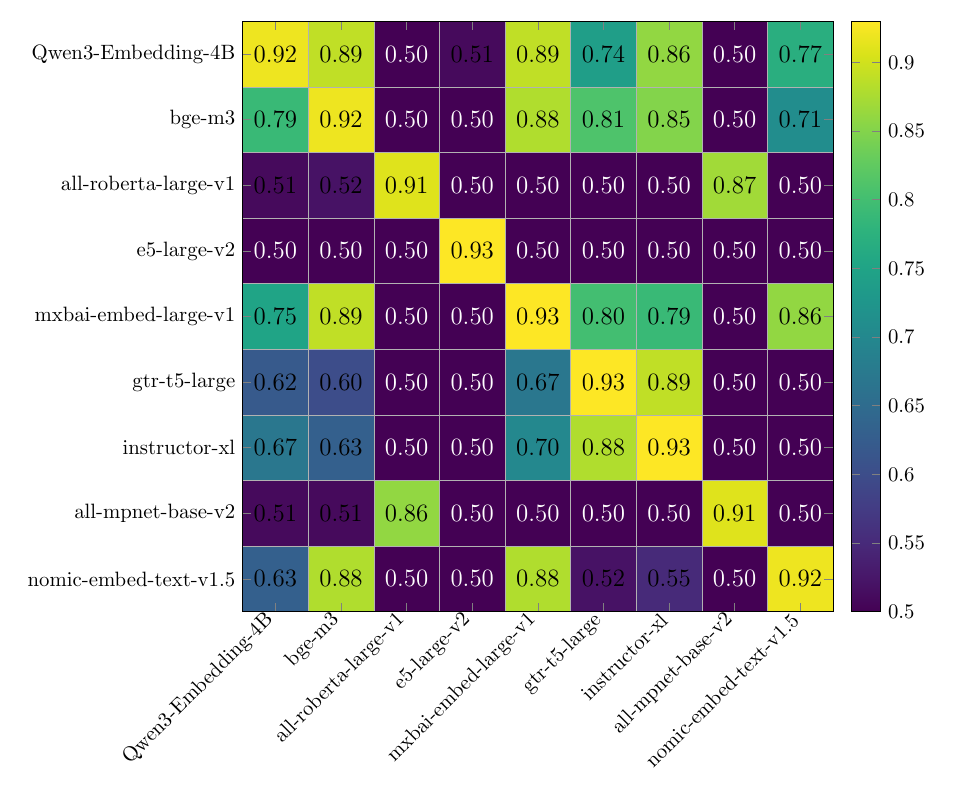}
		\caption*{RR}
	\end{subfigure}%
	\begin{subfigure}[t]{0.22\textwidth}
		\adjincludegraphics[width=\textwidth, trim={{0.25\width} 0 {0.1\width} 0}, clip]{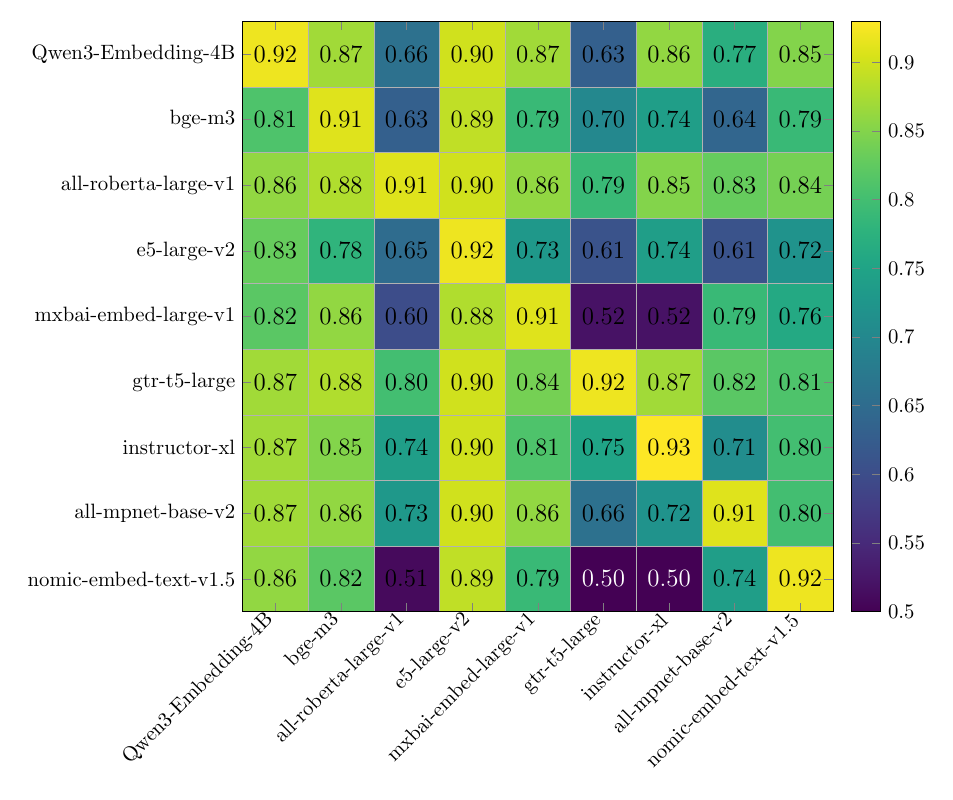}
		\caption*{IRP}
	\end{subfigure}%
	\begin{subfigure}[t]{0.256\textwidth}
		\adjincludegraphics[width=\textwidth, trim={{0.25\width} 0 0 0}, clip]{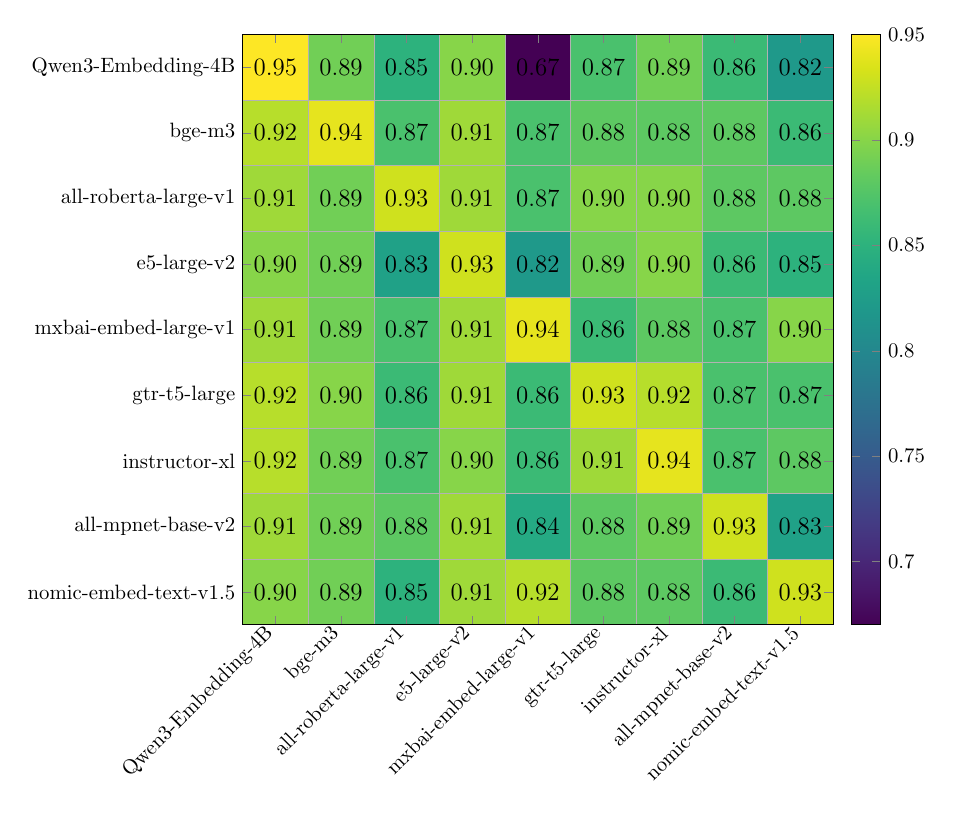}
		\caption*{Linear (SGD)}
	\end{subfigure}%
	\caption{Task-transfer accuracy matrices for SST-2 (rows: source models; columns: target models).}
\end{figure}

\begin{figure}[ht]
	\begin{subfigure}[t]{0.304\textwidth}
		\adjincludegraphics[width=\textwidth, trim={0 0 {0.1\width} 0}, clip]{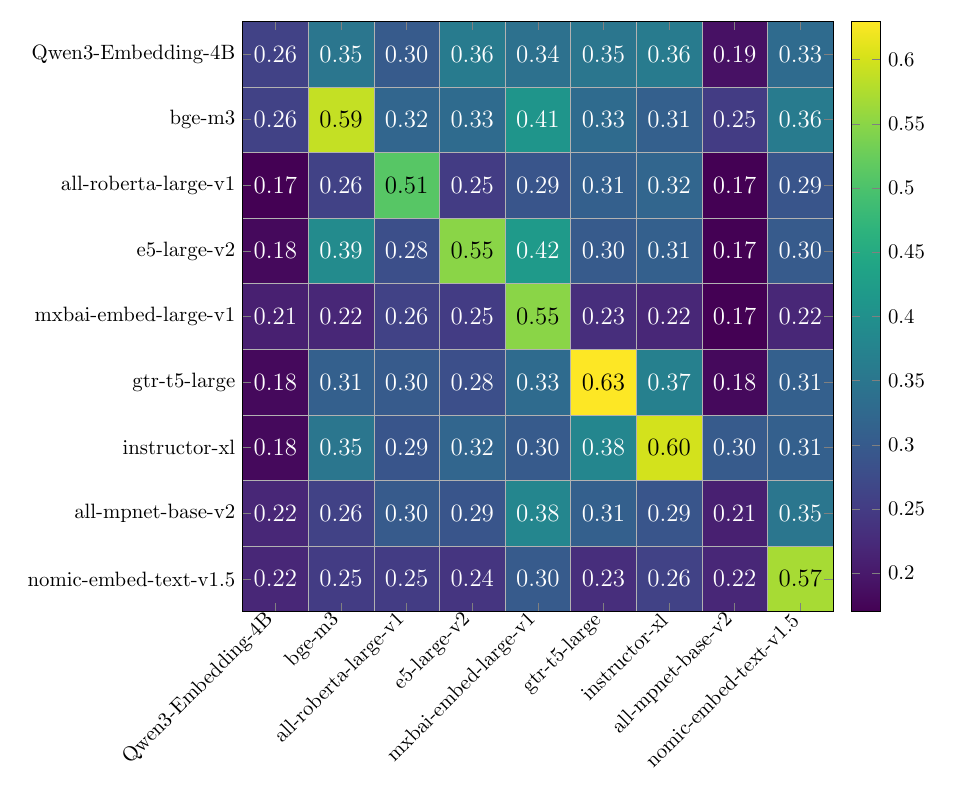}
		\caption*{Linear (pinv)}
	\end{subfigure}%
	\begin{subfigure}[t]{0.22\textwidth}
		\adjincludegraphics[width=\textwidth, trim={{0.25\width} 0 {0.1\width} 0}, clip]{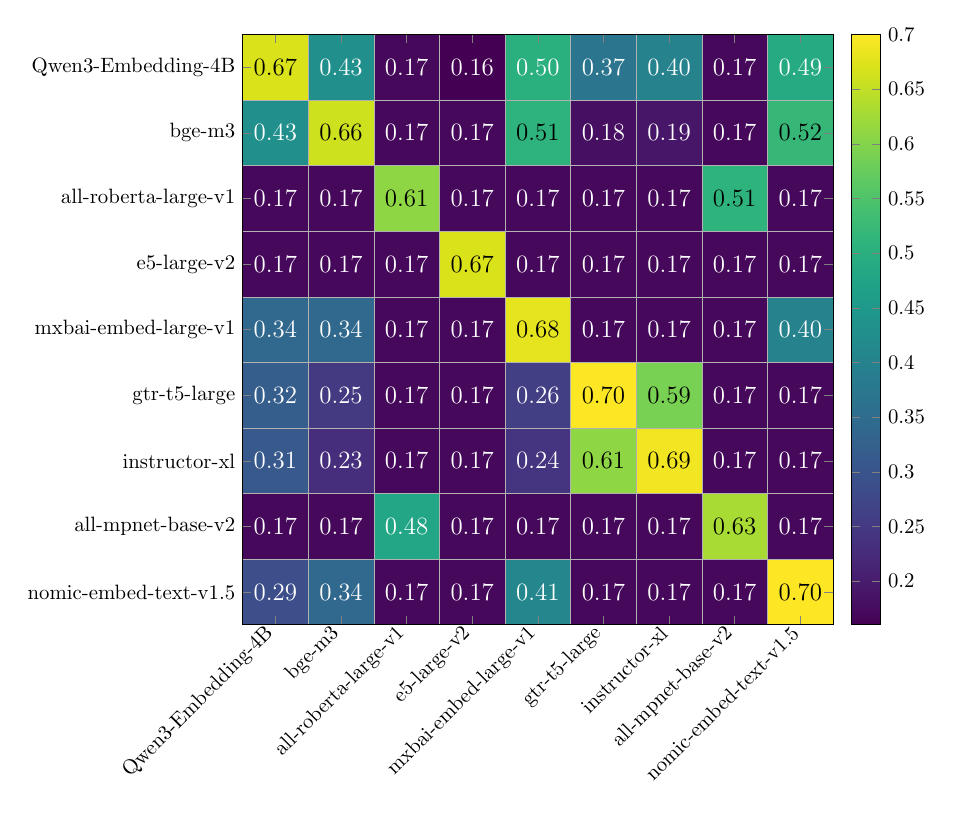}
		\caption*{RR}
	\end{subfigure}%
	\begin{subfigure}[t]{0.22\textwidth}
		\adjincludegraphics[width=\textwidth, trim={{0.25\width} 0 {0.1\width} 0}, clip]{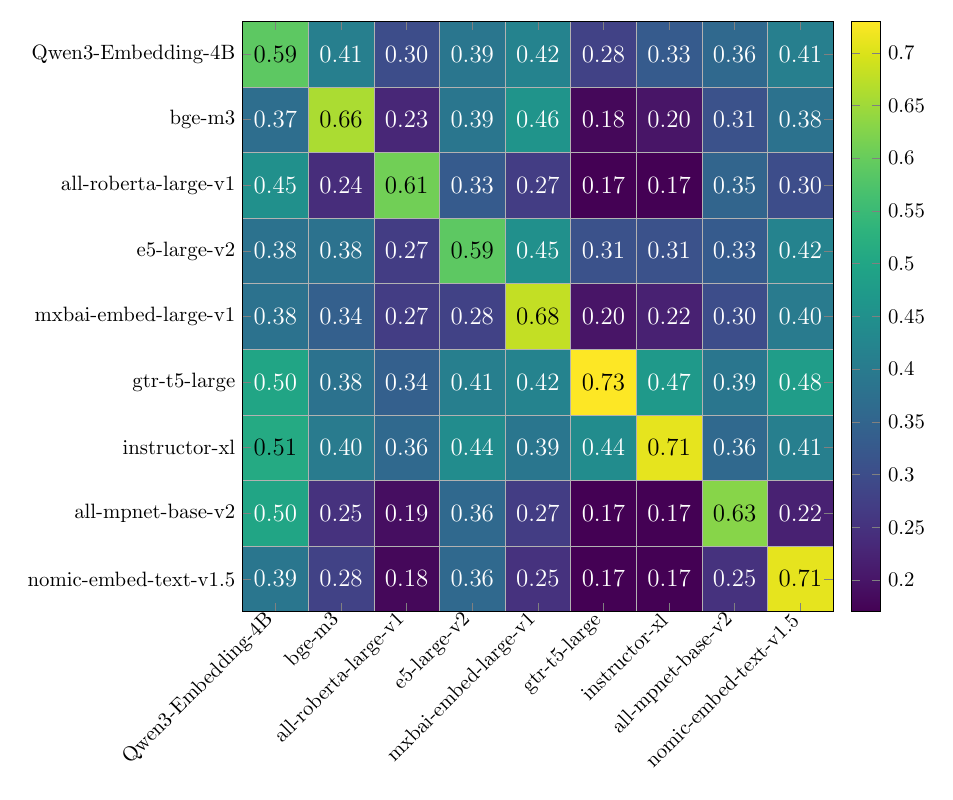}
		\caption*{IRP}
	\end{subfigure}%
	\begin{subfigure}[t]{0.256\textwidth}
		\adjincludegraphics[width=\textwidth, trim={{0.25\width} 0 0 0}, clip]{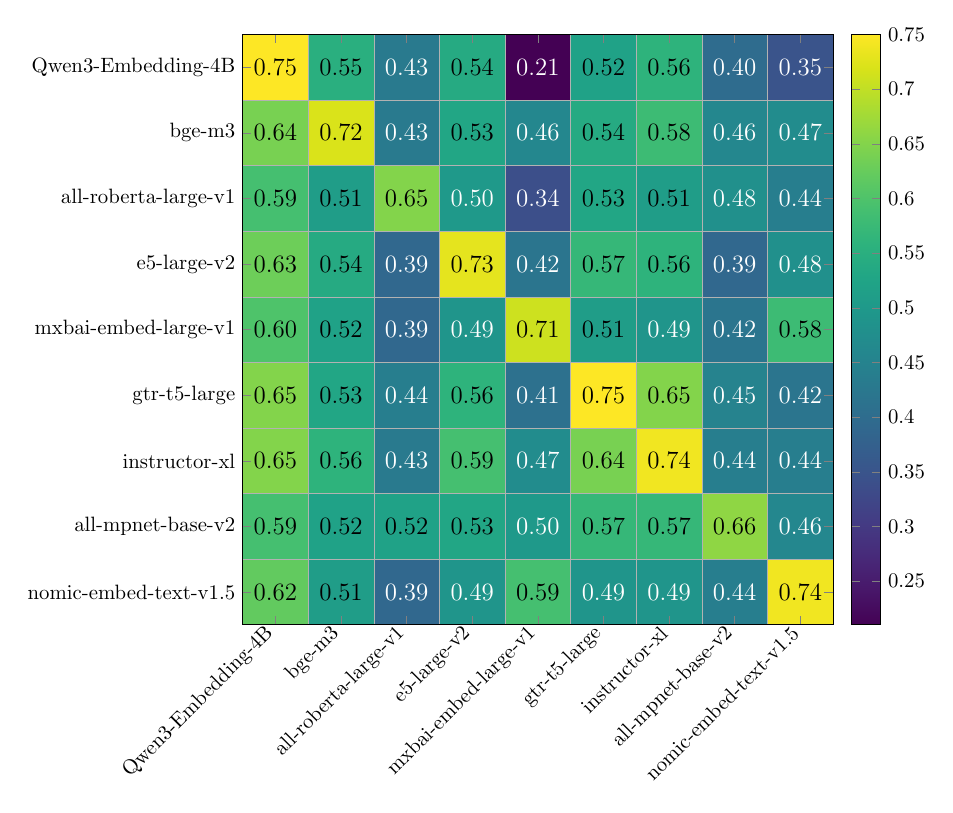}
		\caption*{Linear (SGD)}
	\end{subfigure}%
	\caption{Task-transfer accuracy matrices for Emotion (rows: source models; columns: target models).}
\end{figure}

\begin{figure}[ht]
	\begin{subfigure}[t]{0.304\textwidth}
		\adjincludegraphics[width=\textwidth, trim={0 0 {0.1\width} 0}, clip]{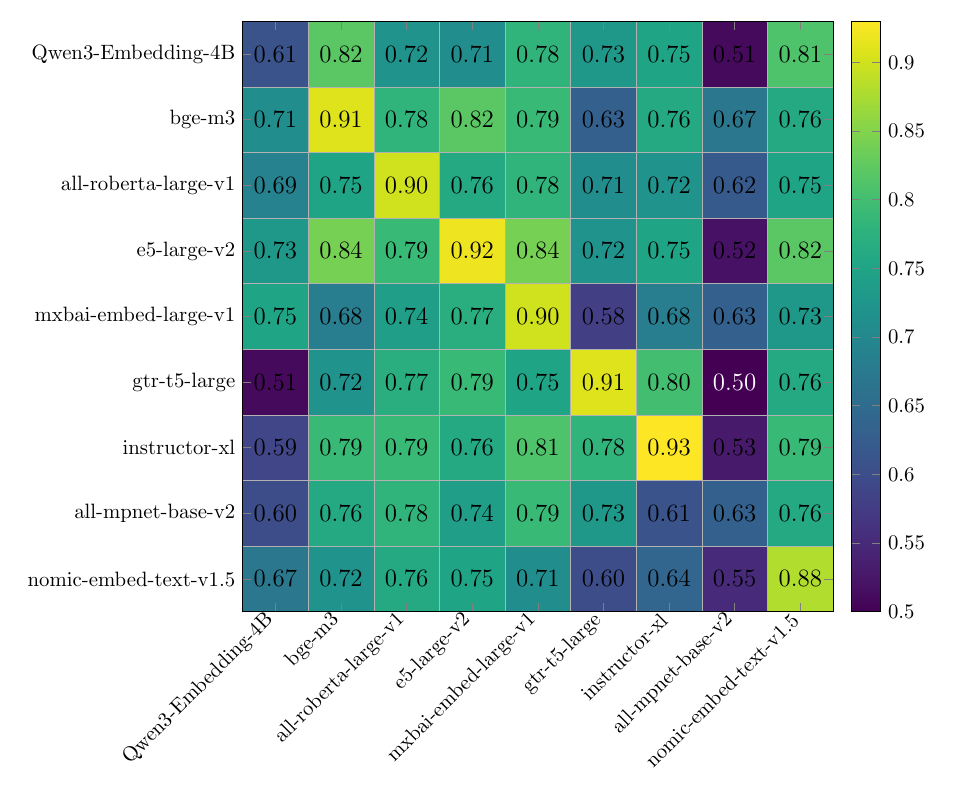}
		\caption*{Linear (pinv)}
	\end{subfigure}%
	\begin{subfigure}[t]{0.22\textwidth}
		\adjincludegraphics[width=\textwidth, trim={{0.25\width} 0 {0.1\width} 0}, clip]{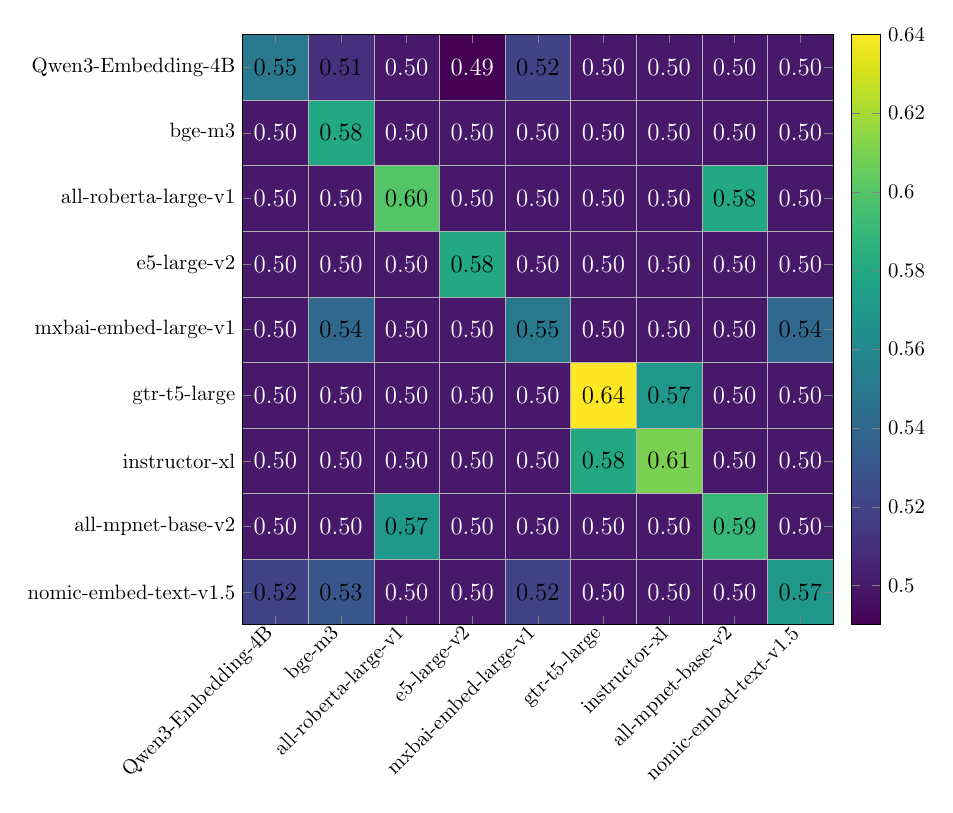}
		\caption*{RR}
	\end{subfigure}%
	\begin{subfigure}[t]{0.22\textwidth}
		\adjincludegraphics[width=\textwidth, trim={{0.25\width} 0 {0.1\width} 0}, clip]{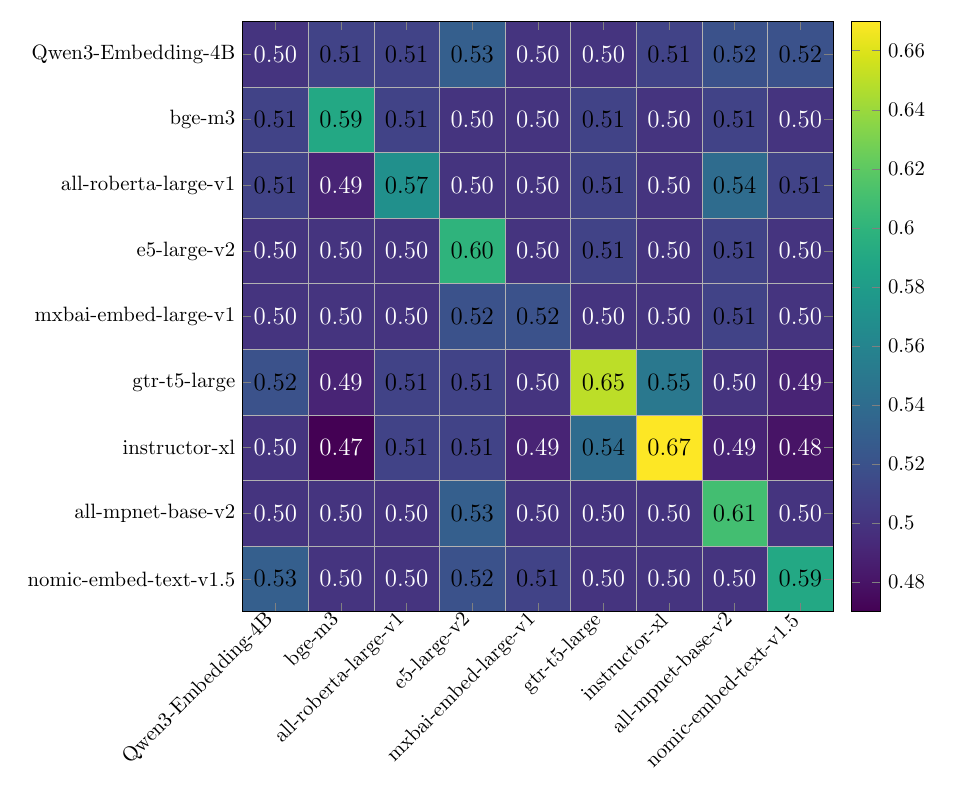}
		\caption*{IRP}
	\end{subfigure}%
	\begin{subfigure}[t]{0.256\textwidth}
		\adjincludegraphics[width=\textwidth, trim={{0.25\width} 0 0 0}, clip]{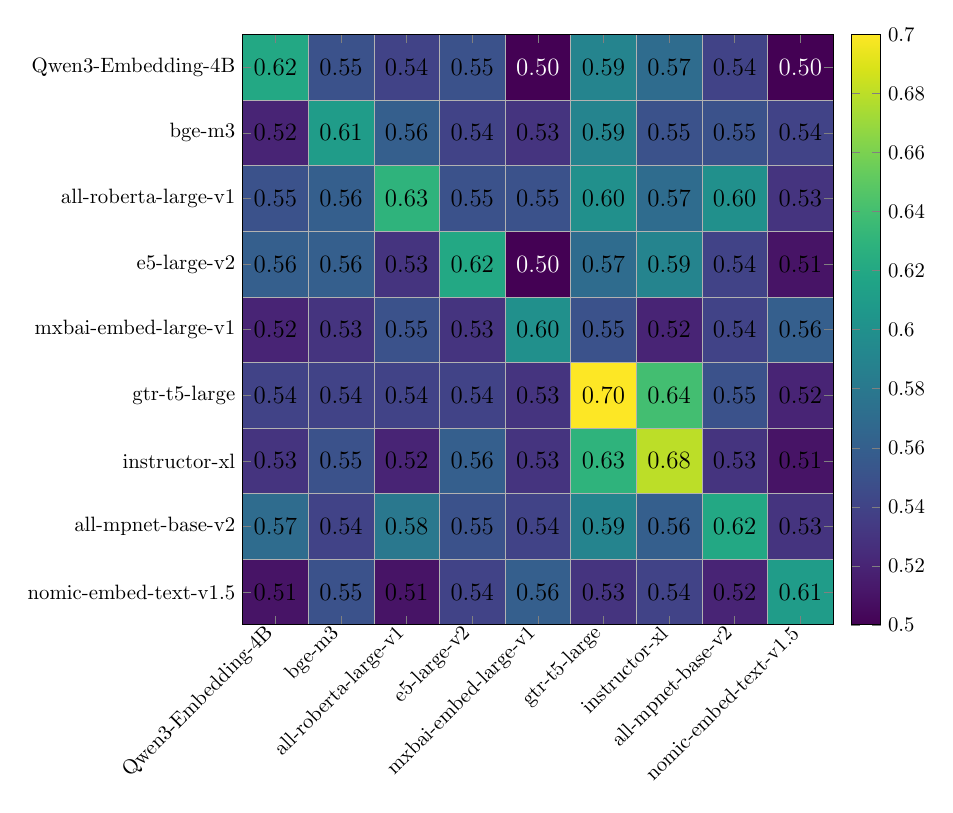}
		\caption*{Linear (SGD)}
	\end{subfigure}%
	\caption{Task-transfer accuracy matrices for Cola (rows: source models; columns: target models).}
\end{figure}

\begin{figure}[ht]
	\begin{subfigure}[t]{0.304\textwidth}
		\adjincludegraphics[width=\textwidth, trim={0 0 {0.1\width} 0}, clip]{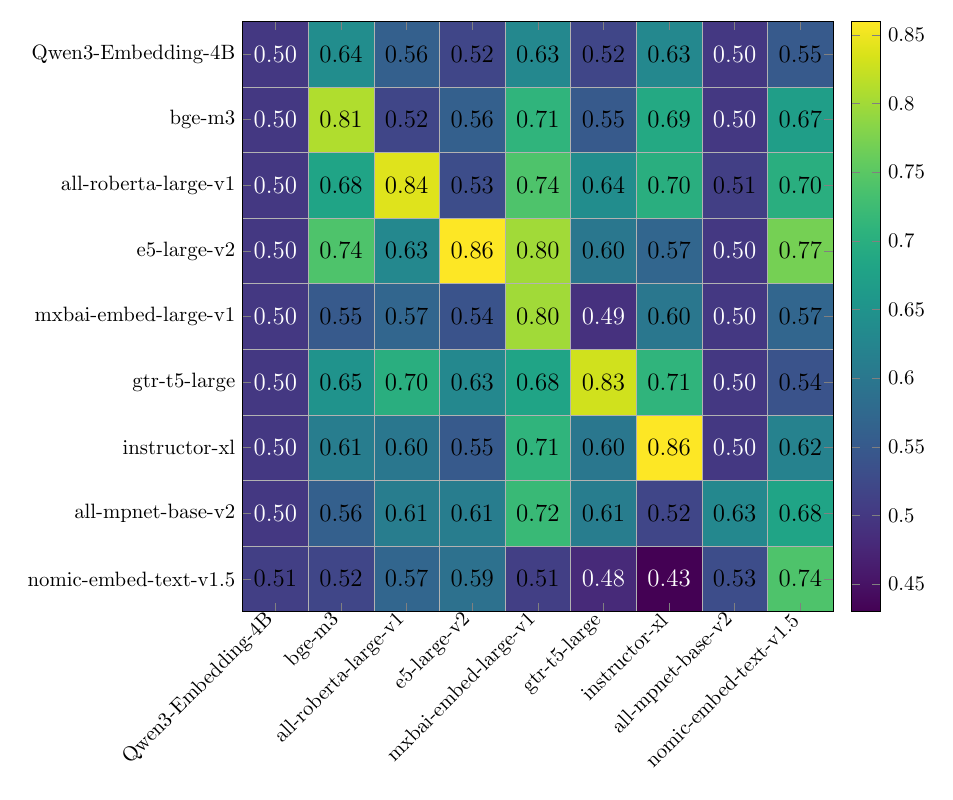}
		\caption*{Linear (pinv)}
	\end{subfigure}%
	\begin{subfigure}[t]{0.22\textwidth}
		\adjincludegraphics[width=\textwidth, trim={{0.25\width} 0 {0.1\width} 0}, clip]{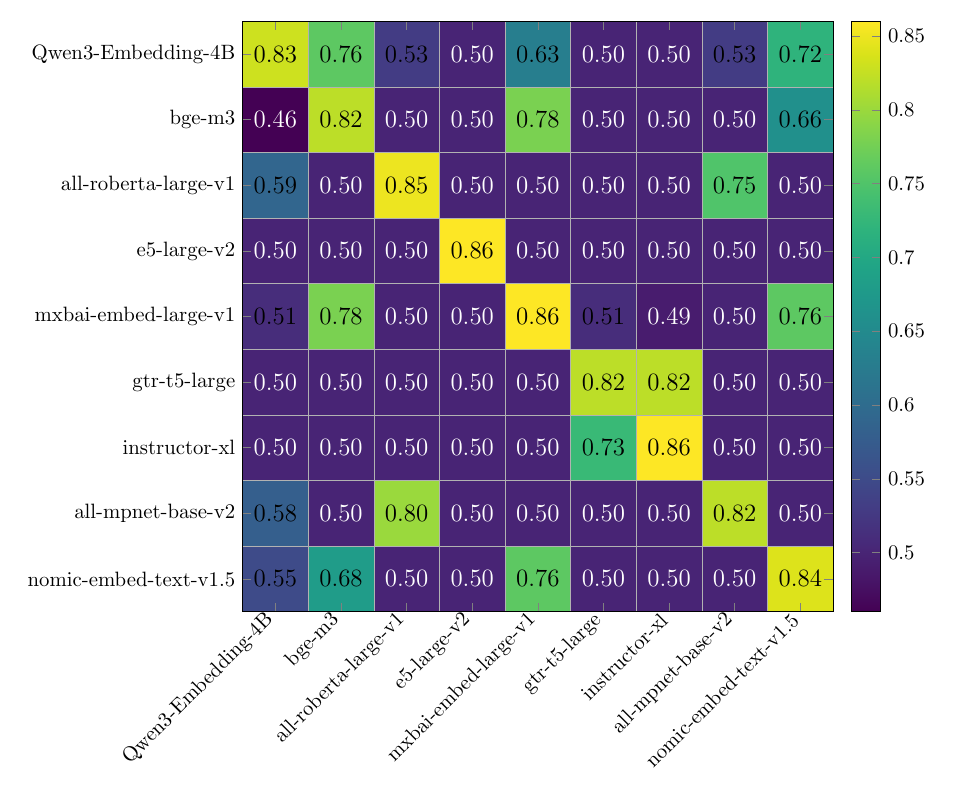}
		\caption*{RR}
	\end{subfigure}%
	\begin{subfigure}[t]{0.22\textwidth}
		\adjincludegraphics[width=\textwidth, trim={{0.25\width} 0 {0.1\width} 0}, clip]{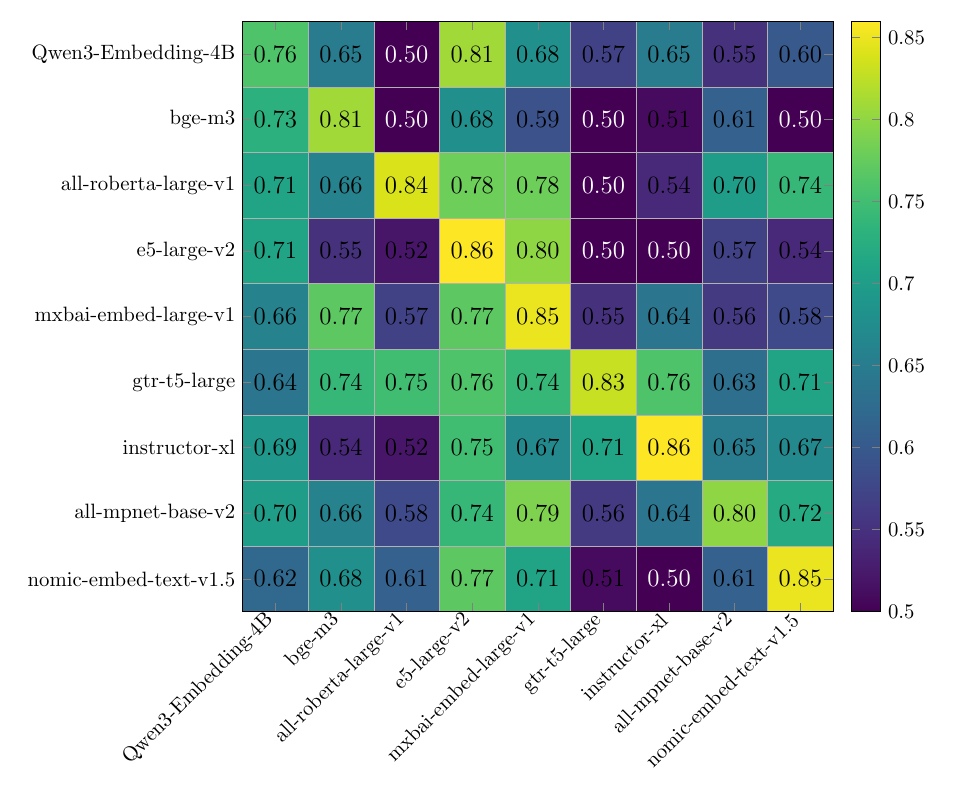}
		\caption*{IRP}
	\end{subfigure}%
	\begin{subfigure}[t]{0.256\textwidth}
		\adjincludegraphics[width=\textwidth, trim={{0.25\width} 0 0 0}, clip]{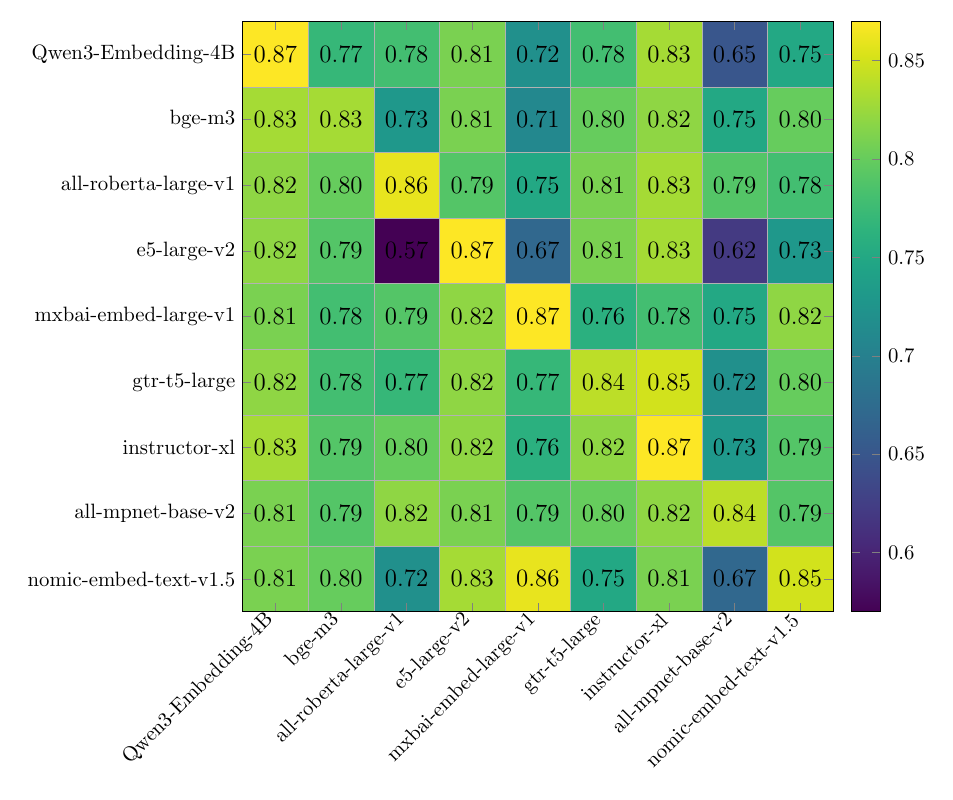}
		\caption*{Linear (SGD)}
	\end{subfigure}%
	\caption{Task-transfer accuracy matrices for IMDB (rows: source models; columns: target models).}
\end{figure}

\begin{figure}[ht]
	\begin{subfigure}[t]{0.304\textwidth}
		\adjincludegraphics[width=\textwidth, trim={0 0 {0.1\width} 0}, clip]{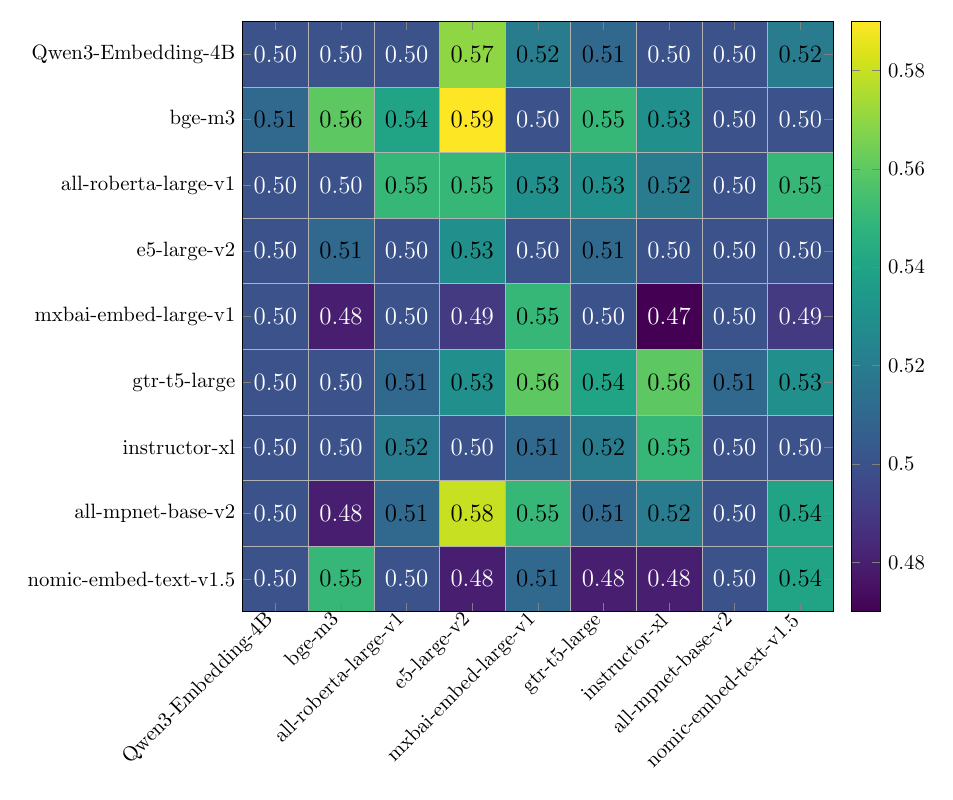}
		\caption*{Linear (pinv)}
	\end{subfigure}%
	\begin{subfigure}[t]{0.22\textwidth}
		\adjincludegraphics[width=\textwidth, trim={{0.25\width} 0 {0.1\width} 0}, clip]{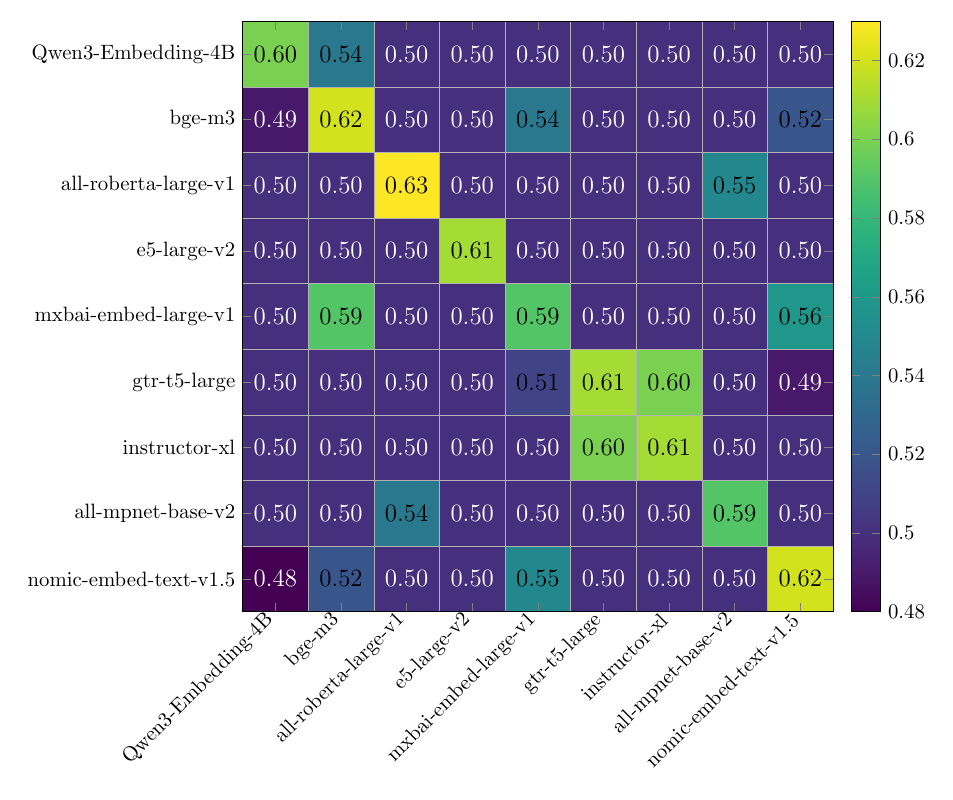}
		\caption*{RR}
	\end{subfigure}%
	\begin{subfigure}[t]{0.22\textwidth}
		\adjincludegraphics[width=\textwidth, trim={{0.25\width} 0 {0.1\width} 0}, clip]{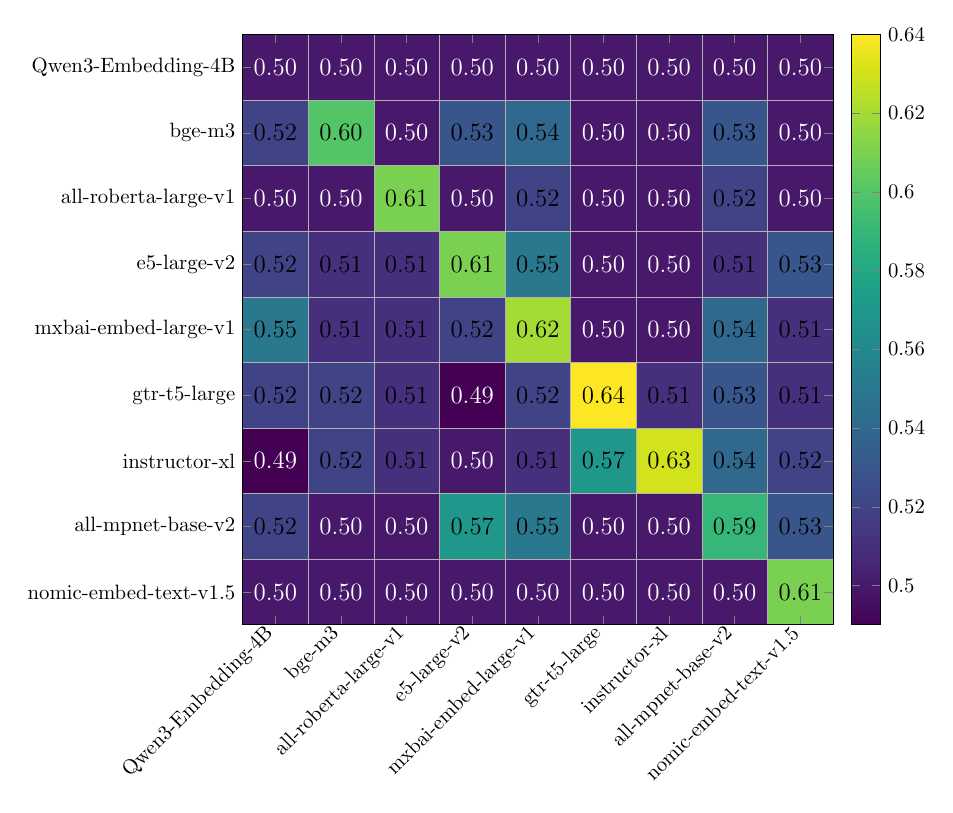}
		\caption*{IRP}
	\end{subfigure}%
	\begin{subfigure}[t]{0.256\textwidth}
		\adjincludegraphics[width=\textwidth, trim={{0.25\width} 0 0 0}, clip]{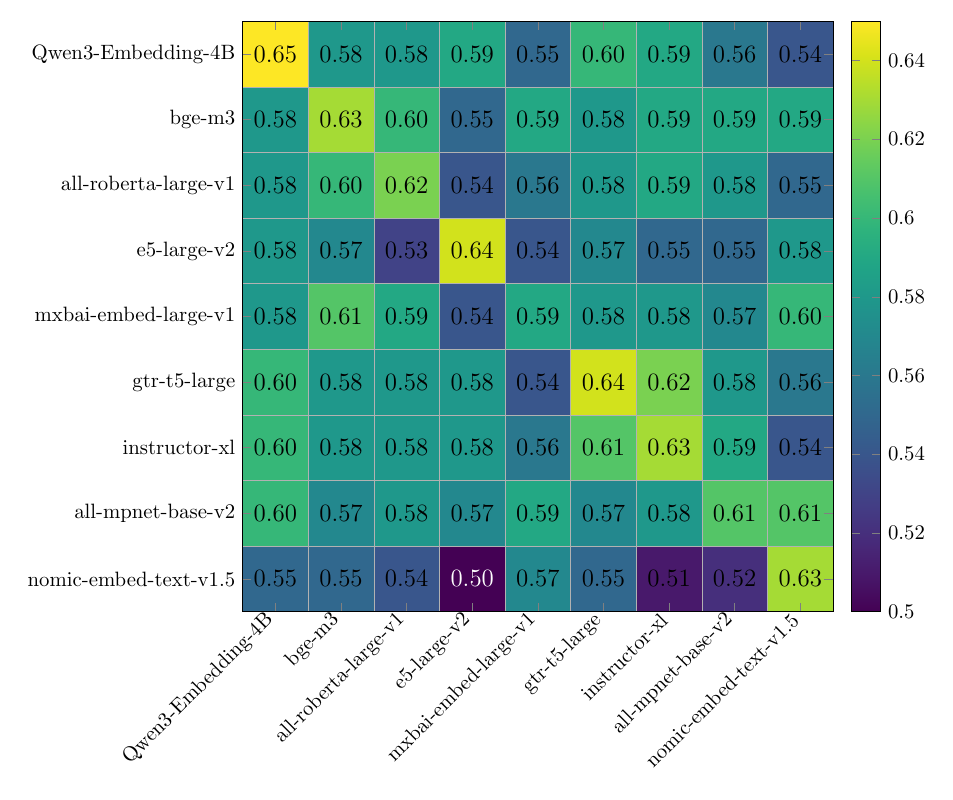}
		\caption*{Linear (SGD)}
	\end{subfigure}%
	\caption{Task-transfer accuracy matrices for MRPC (rows: source models; columns: target models).}
\end{figure}

\begin{figure}[ht]
	\begin{subfigure}[t]{0.304\textwidth}
		\adjincludegraphics[width=\textwidth, trim={0 0 {0.1\width} 0}, clip]{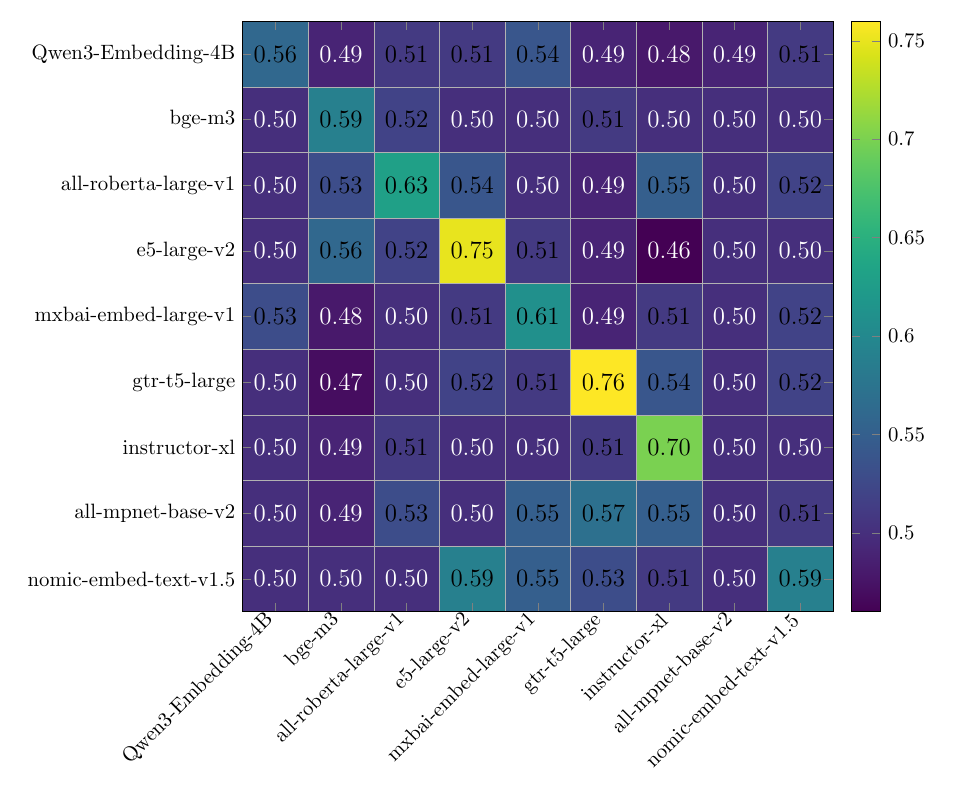}
		\caption*{Linear (pinv)}
	\end{subfigure}%
	\begin{subfigure}[t]{0.22\textwidth}
		\adjincludegraphics[width=\textwidth, trim={{0.25\width} 0 {0.1\width} 0}, clip]{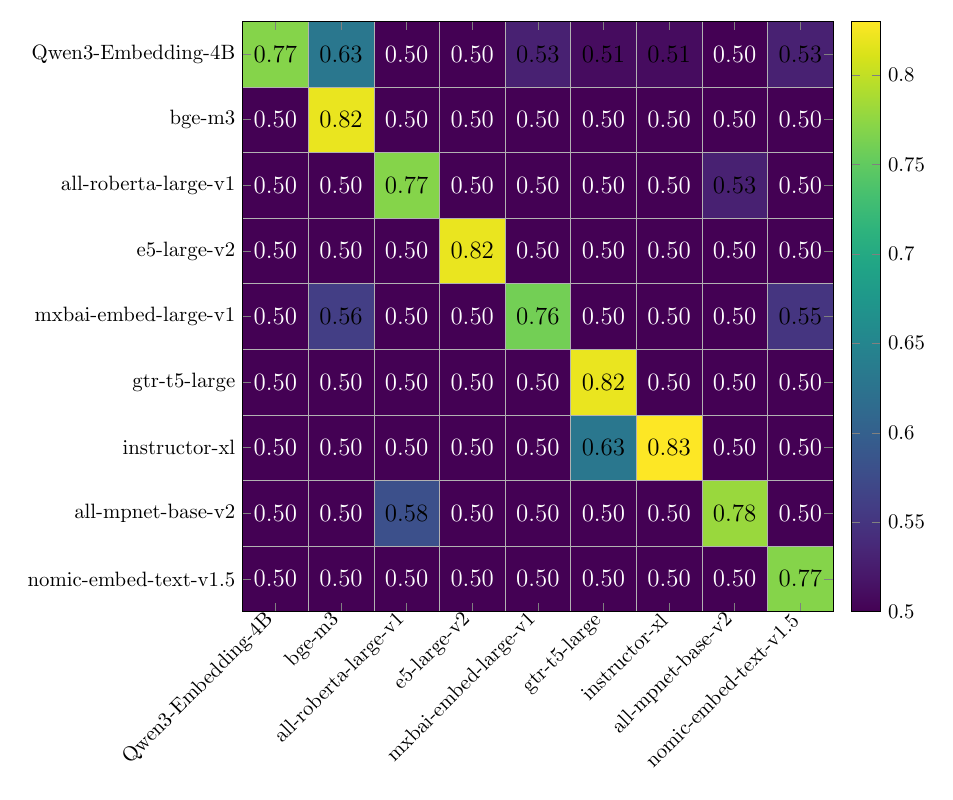}
		\caption*{RR}
	\end{subfigure}%
	\begin{subfigure}[t]{0.22\textwidth}
		\adjincludegraphics[width=\textwidth, trim={{0.25\width} 0 {0.1\width} 0}, clip]{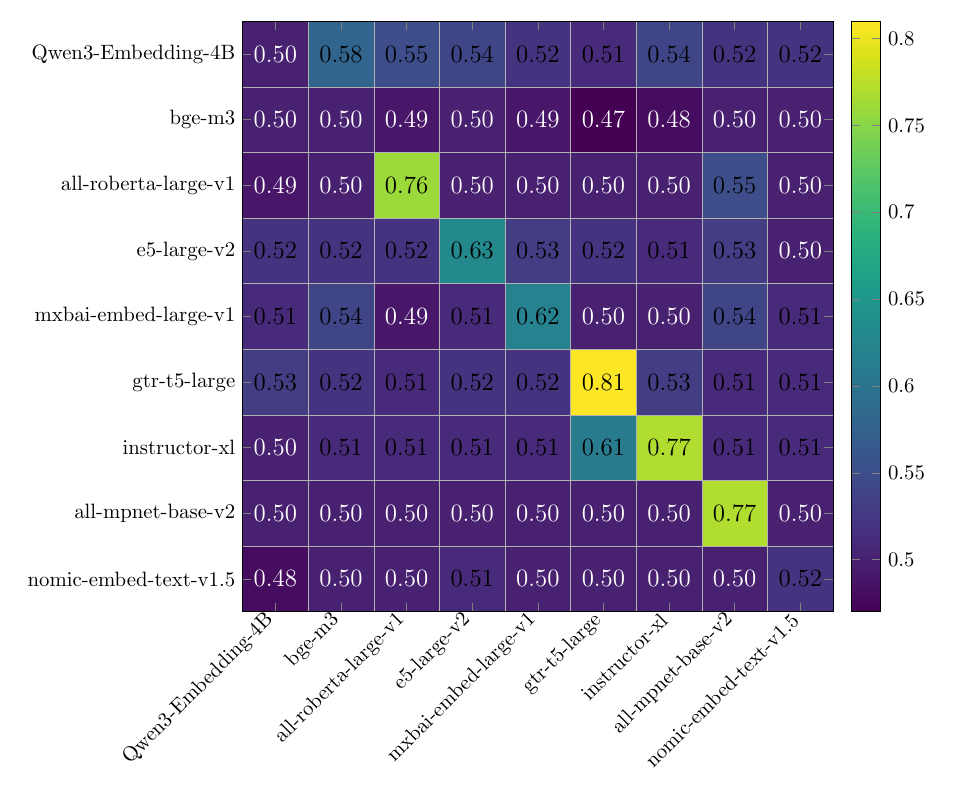}
		\caption*{IRP}
	\end{subfigure}%
	\begin{subfigure}[t]{0.256\textwidth}
		\adjincludegraphics[width=\textwidth, trim={{0.25\width} 0 0 0}, clip]{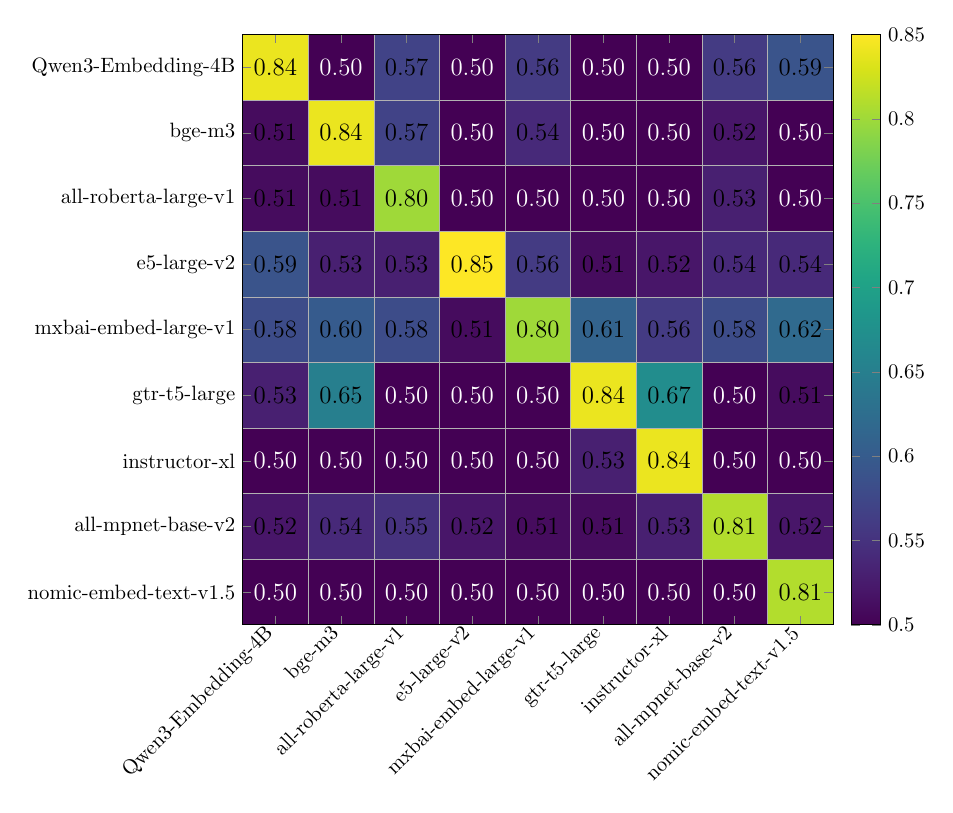}
		\caption*{Linear (SGD)}
	\end{subfigure}%
	\caption{Task-transfer accuracy matrices for QQP (rows: source models; columns: target models).}
\end{figure}

\section{Hubness analysis}
\label{app:hubness}
Tables~\ref{tab:hubness_stats_top_1}, \ref{tab:hubness_stats_top_5}, and \ref{tab:hubness_stats_top_10} report hubness statistics for the embeddings of 1000 of the most frequent English words. The tables show how many times each word appears as a Top-1/5/10 neighbor of other words for each model. We compute the 95th percentile, which indicates high hubness, and the 10th percentile, which indicates antihubness if low (points that are rarely retrieved as neighbors). We observe that the models with high hubness (high 95th percentile, high skewness, and high variance) are the least performing models as targets for retrieval in $k$-NN. In particular, \texttt{e5-large-v2} has one of the highest hubness indications, and also one of the highest antihubness indications (low 10th percentile and low median), this is a strong indication of why \texttt{e5-large-v2} performs poorly in $k$-NN retrieval. Curiously, \texttt{Qwen3-Embedding-4B} also has a high hubness indication, which is consistent with its performance in $k$-NN as a target model. However, since it did well as a source model, this suggests that hubness cannot explain everything about cross-model compatibility.

\begin{table}[ht]
	\begin{tabular}{lcccccc}
		\hline
		Model                 & Percentil 95 & Percentil 10 & Max & Median & Skewness & Variance \\
		\hline
		Qwen3-Embedding-4B    & 1.00         & 1.00         & 2.0 & 1.0    & 0.00     & 0.03     \\
		all-roberta-large-v1  & 1.00         & 1.00         & 3.0 & 1.0    & 3.23     & 0.02     \\
		e5-large-v2           & 1.00         & 1.00         & 5.0 & 1.0    & 9.66     & 0.04     \\
		bge-m3                & 1.00         & 1.00         & 1.0 & 1.0    & 0.00     & 0.00     \\
		mxbai-embed-large-v1  & 1.00         & 1.00         & 1.0 & 1.0    & 0.00     & 0.00     \\
		gtr-t5-large          & 1.00         & 1.00         & 1.0 & 1.0    & 0.00     & 0.00     \\
		all-mpnet-base-v2     & 1.00         & 1.00         & 1.0 & 1.0    & 0.00     & 0.00     \\
		instructor-xl         & 1.00         & 1.00         & 1.0 & 1.0    & 0.00     & 0.00     \\
		nomic-embed-text-v1.5 & 1.00         & 1.00         & 1.0 & 1.0    & 0.00     & 0.00     \\
		\hline
	\end{tabular}
	\caption{Hubness Statistics for Top 1}\label{tab:hubness_stats_top_1}
\end{table}

\begin{table}[ht]
	\begin{tabular}{lcccccc}
		\hline
		Model                 & Percentil 95 & Percentil 10 & Max   & Median & Skewness & Variance \\
		\hline
		Qwen3-Embedding-4B    & 14.05        & 1.00         & 125.0 & 3.0    & 7.36     & 51.24    \\
		all-roberta-large-v1  & 11.00        & 2.00         & 108.0 & 4.0    & 10.07    & 29.58    \\
		e5-large-v2           & 18.05        & 1.00         & 181.0 & 2.0    & 7.64     & 148.64   \\
		bge-m3                & 12.00        & 1.00         & 38.0  & 4.0    & 2.88     & 19.22    \\
		mxbai-embed-large-v1  & 10.00        & 1.00         & 329.0 & 4.0    & 18.75    & 161.95   \\
		gtr-t5-large          & 11.00        & 2.00         & 33.0  & 4.0    & 2.16     & 11.12    \\
		all-mpnet-base-v2     & 10.00        & 2.00         & 23.0  & 4.0    & 1.30     & 8.50     \\
		instructor-xl         & 12.00        & 2.00         & 33.0  & 4.0    & 2.04     & 12.27    \\
		nomic-embed-text-v1.5 & 10.00        & 2.00         & 111.0 & 4.0    & 13.43    & 20.39    \\
		\hline
	\end{tabular}
	\caption{Hubness Statistics for Top 5}\label{tab:hubness_stats_top_5}
\end{table}

\begin{table}[ht]
	\begin{tabular}{lcccccc}
		\hline
		Model                 & Percentil 95 & Percentil 10 & Max   & Median & Skewness & Variance \\
		\hline
		Qwen3-Embedding-4B    & 31.05        & 2.00         & 318.0 & 5.0    & 8.41     & 311.17   \\
		all-roberta-large-v1  & 22.00        & 3.00         & 224.0 & 7.0    & 8.93     & 191.79   \\
		e5-large-v2           & 46.00        & 1.00         & 324.0 & 3.0    & 6.80     & 702.73   \\
		bge-m3                & 27.00        & 2.00         & 124.0 & 7.0    & 4.05     & 123.09   \\
		mxbai-embed-large-v1  & 20.05        & 2.00         & 683.0 & 6.0    & 13.70    & 1025.52  \\
		gtr-t5-large          & 23.00        & 3.00         & 90.0  & 8.0    & 3.41     & 59.63    \\
		all-mpnet-base-v2     & 20.00        & 4.00         & 41.0  & 9.0    & 1.46     & 33.85    \\
		instructor-xl         & 24.00        & 3.00         & 102.0 & 8.0    & 3.22     & 68.72    \\
		nomic-embed-text-v1.5 & 22.00        & 3.00         & 371.0 & 8.0    & 19.95    & 179.15   \\
		\hline
	\end{tabular}
	\caption{Hubness Statistics for Top 10}\label{tab:hubness_stats_top_10}
\end{table}

\clearpage
\newpage

\end{document}